\documentclass[a4paper,fleqn]{cas-dc}

\usepackage[numbers]{natbib}
\usepackage{amssymb}
\usepackage{amsmath,amsfonts}
\usepackage{array}
\usepackage[caption=false,font=normalsize,labelfont=sf,textfont=sf]{subfig}
\usepackage{textcomp}
\usepackage{url}
\usepackage{verbatim}
\usepackage[dvipsnames]{xcolor}
\definecolor{assumpH1}{RGB}{237, 125, 49}
\definecolor{assumpH2}{RGB}{113, 174, 71}
\definecolor{assumpH3}{RGB}{0, 176, 240}
\definecolor{assumpH41}{RGB}{112, 48, 160}
\definecolor{assumpH42}{RGB}{255, 0, 0}
\definecolor{assumpH5}{RGB}{255, 51, 204}
\usepackage{pgfplots}
\pgfplotsset{width=7cm,compat=1.8}
\usepackage{pgfplotstable}
\usepackage{multirow}
\usepackage{floatrow}

\def\tsc#1{\csdef{#1}{\textsc{\lowercase{#1}}\xspace}}
\tsc{WGM}
\tsc{QE}
\tsc{EP}
\tsc{PMS}
\tsc{BEC}
\tsc{DE}
\begin{document}
\let\WriteBookmarks\relax
\def\floatpagepagefraction{1}
\def\textpagefraction{.001}

% Short title
\shorttitle{Deep state-space modeling for explainable human movement representation}

% Short author
\shortauthors{B. E. Olivas-Padilla et~al.}

% Main title of the paper
\title [mode = title]{Deep state-space modeling for explainable representation, analysis, and generation of professional human poses}   
\tnotemark[1]
% Title footnote mark
% eg: \tnotemark[1]
% \tnotemark[1,2]

% Title footnote 1.
% eg: \tnotetext[1]{Title footnote text}
% \tnotetext[<tnote number>]{<tnote text>} 
\tnotetext[1]{The research leading to these results has received funding by the~EU Horizon 2020 Research and Innovation Programme under Grant Agreement No. 820767, CoLLaboratE project, and Grant No. 822336, Mingei project.}

\author[mines]{Brenda Elizabeth Olivas-Padilla}
\cormark[1]
\credit{Methodology, Data curation, Writing - Original draft preparation, Software, Validation}
\ead{brenda.olivas@minesparis.psl.eu}

\author[mines]{Alina Glushkova}
\credit{Formal analysis, Review - Original draft preparation, Funding acquisition}
\ead{alina.glushkova@minesparis.psl.eu}

\author[mines]{Sotiris Manitsaris}
\credit{Conceptualization of this study, Methodology, Review - Original draft preparation, Funding acquisition}
\ead{sotiris.manitsaris@minesparis.psl.eu}

\affiliation[mines]{organization={Centre for Robotics, Mines Paris, Université PSL},%Department and Organization
            addressline={}, 
            city={Paris},
            postcode={75006}, 
            state={},
            country={France}}

\cortext[cor1]{Corresponding author}

% Here goes the abstract
\begin{abstract}
The analysis of human movements has been extensively studied due to its wide variety of practical applications, such as human-robot interaction, human learning applications, or clinical diagnosis. Nevertheless, the state-of-the-art still faces scientific challenges when modeling human movements. To begin, new models must account for the stochasticity of human movement and the physical structure of the human body in order to accurately predict the evolution of full-body motion descriptors over time. Second, while utilizing deep learning algorithms, their explainability in terms of body posture predictions needs to be improved as they lack comprehensible representations of human movement. This paper addresses these challenges by introducing three novel methods for creating explainable representations of human movement. In this study, human body movement is formulated as a state-space model adhering to the structure of the Gesture Operational Model (GOM), whose parameters are estimated through the application of deep learning and statistical algorithms. The trained models are used for the full-body dexterity analysis of expert professionals, in which dynamic associations between body joints are identified, and for generating artificially professional movements.
\end{abstract}

% Use if graphical abstract is present
% \begin{graphicalabstract}
% \includegraphics{figs/grabs.pdf}
% \end{graphicalabstract}

% Research highlights
\begin{highlights}
\item Three novel methods for creating explainable mathematical representations of human movement.
\item Analysis of full-body dexterity in industrial operators and expert artisans.
\item Generation of professional movements utilizing explainable models trained with either one-shot learning or deep learning.
\end{highlights}

% Keywords
% Each keyword is seperated by \sep

\maketitle

\section{Introduction}
Movement is an essential component of human life. Humans continually exchange information and interact with their surroundings through their movements, which result from complex and highly coordinated mechanical interaction between bones, muscles, ligaments, and joints within the musculoskeletal system. Through the study of this interaction and its effects, the structure, function, and motion of human bodies can be examined, and the resulting knowledge can be used to improve the quality of human life. Recently, the automatic analysis of human movements has gained significance as a research domain. This is due to the emergence of various applications in health, sports, motion-driven user interfaces, and intelligent surveillance that utilize motion capture (MoCap) data. Depending on the motion-related application, various techniques, such as statistical models or deep-learning approaches,  have been used to model human motion data. However, it remains complex and requires overcoming scientific challenges to design an accurate and versatile automatic analysis tool to describe human motion dynamics based on MoCap data. A first challenge would be achieving a high level of modeling accuracy. The system should consider human motion's stochasticity and body joints' mediations to accurately predict how human motion descriptors will evolve over time so that this information can be used proactively. Secondly, the models should be able to explain their outputs and assist humans in understanding how specific human movements are performed in order to properly employ this knowledge in an application, for instance, in a human learning application.

This paper addresses the previous challenges through the analytical modeling of human motion dynamics for creating explainable models of human movements. Accordingly, the following hypothesis was formulated: Analytical modeling of human motion dynamics is possible by incorporating meaningful assumptions about interdependencies between joints as well as dependencies between their prior values. To address this hypothesis, it was first investigated statistical and data-driven approaches for parameterizing interpretable mathematical representations of human movements. The models are then trained with professional movements to determine the contribution and significance of the spatial (body joints trajectories) and temporal (previous states) assumptions incorporated into the model. Finally, these results are used for body dexterity analysis, where the parameters of the models provide insight into how skilled artisans and operators use their whole bodies when performing their professional tasks. This analysis can be used to determine which body joints are most meaningful in modeling professional tasks, and trained models can also be utilized to generate artificial human movements.

This article first reviews relevant research in human motion modeling in Section \ref{sec:soa}. Section \ref{sec:method} then introduces the three novel approaches for modeling human movements through explainable mathematical representations based on GOM. The application of trained motion representations for body dexterity analysis is described in Section \ref{sec:bod}. Next, Section \ref{sec:sim} presents and discusses the human movement generation performance with each dataset used. Last, Section \ref{sec:concl} provides the conclusions of this work.

\section{Related Works}\label{sec:soa}
This section presents diverse methodologies for the modeling and analysis of full-body human movements. These are organized in accordance with the purpose of their analysis, and it is discussed how they represent human movement and their current limitations.

\subsection{Biomechanical modeling}
Biomechanical models have been applied to simulate human movement and analyze the changes that occur as a result of internal and external action forces. These models represent the human body as a set of articulated links in a kinetic chain, with joint torques and forces calculated using anthropometric, postural, and hand load data \cite{Muller2020}. Inverse dynamics is applied to extract quantitative information about the mechanics of the musculoskeletal system during the performance of a motor task. Previous research has employed biomechanical modeling to extract the joint's kinematic and kinetic contributions to a variety of motor tasks and then examined the joint's mechanical loading and reaction for either ergonomic interventions, sports, or medical research. Menychtas et al. \cite{Menychtas2020} used the Newton-Euler algorithm to compute upper body joint torques in order to study the ergonomic impact of various positions on human joints. Wang et al. \cite{Wang2019} and Gandhi et al. \cite{Gandhi2019} used finite element models to investigate the mechanics of spinal cord injuries. Similarly, Cazzola et al. \cite{Cazzola2017} used a finite element model to analyze the forces and loads experienced by the cervical spine during rugby scrummaging.

Developing precise and noninvasive methods in biomechanical modeling remains challenging. The need for new methods to analyze motions captured outside of laboratories arises as laboratory recordings lack authenticity. They are not done on the actual scenarios where the movements are performed, presumably leading to inaccurate measurements. Moreover, they typically assume deterministic and well-defined movements and do not account for variability and uncertainty in human movement. In contrast, stochastic models or hybrid biomechanical-stochastic models have been used to model variability and uncertainty to analyze and generate a range of plausible movements.

\subsection{Stochastic modeling}
Stochastic modeling has been used to learn the unpredictable nature of human motion. These models utilize the variance information contained in body motion trajectories to predict and identify human intentions and actions. Among the most successful methods for dealing with the temporal variations of human movements are generative models, in which time series are reorganized by sequential states. A common approach to describing human motion in this way is by using state-space models (SSMs). Previous research applied SSMs based on Kalman filters for representing kinematic models to forecast pedestrian position trajectories \cite{Barth2008, Binelli2005}. The Kalman filter (KF) was mostly used to track the position of pedestrians based on their estimated velocity or acceleration. Caramiaux et al. \cite{Caramiaux2015} introduced an adaptive SSM based on particle filtering that recognizes and continuously tracks the variations of gestures. Hidden Markov Models (HMMs) have proven successful in modeling the temporal evolution of gestures globally \cite{Glushkova2018, Malaise2018, Manitsaris2020}. Other stochastic models used to model full-body movements include the Dynamic Naive Bayes model proposed by Devanne et al. \cite{Devanne2017}. The model captured the dynamics of motion primitives and segmented continuously distinct human behaviors in extended sequences. In a separate study, Ravichandar et al. \cite{Ravichandar2019} proposed the Intention-Driven Dynamics Model that presupposes human dynamics change when actions are motivated by various intentions. 

Despite these encouraging advances in recent years, accurate human motion modeling in unconstrained environments remains challenging. Unresolved difficulties in the modeling concern spatiotemporal dynamics. For example, even the same motion done by the same individual can have varying speeds and starting/ending positions, let alone in scenarios involving multiple performers. Hybrid biomechanical-stochastic models have been developed in an effort to overcome the aforementioned challenges and improve the modeling performance of simple stochastic models. Besides capturing human motion stochastics, these also encapsulate the kinematic correlations or dependencies among different skeletal joints. 

\subsection{Hybrid biomechanical-stochastic modeling}
The analysis of the random outcomes of human movement has been improved by developing hybrid methodologies that consider both the human biomechanical structure and the stochastic nature of human motion \cite{Shi2003, Lin2012, Bologna2020}. This type of model has been extensively used to study musculoskeletal pathologies. By human motion modeling, the deviations from normal movement in terms of altered kinematic or kinetic patterns are identified and then utilized to evaluate neuromusculoskeletal conditions, to aid in subsequent treatment planning, or to analyze the efficacy of treatment in different patient groups. A hybrid model designed to predict the probability of injury and identify factors contributing to the risk of non-contact anterior cruciate ligament (ACL) injuries, has been proposed by Lin et al. \cite{Lin2012}. A biomechanical model of the ACL estimated the lower leg kinematics and kinetics. In turn, the means and standard deviations of the number of simulated non-contact ACL injuries, injury rate and female-to-male injury rate, were calculated in Monte Carlo simulations of non-contact ACL injury and non-injury trials. In another work, Donnell et al. \cite{Donnell2014} used a two-state Markov chain model to represent the survival of surgical repair from a torn rotator cuff. The load applied to the shoulder and the structural capacity of tissue were defined as the random variables.

A limitation of hybrid models, as well as biomechanical and stochastic models, is that their processing requirements grow exponentially as the number of model parameters increases. As a result, they are not practical for analyzing large datasets of human movements or high-dimensional MoCap data as deep learning approaches (described in the next section), requiring feature selection or extraction algorithms. Moreover, to adequately design the hybrid models and their assumptions, it is necessary to have prior knowledge of the data that would be used for the training. For example, if the objective is to understand muscle coordination, a model that omits joints and muscles is unlikely to be helpful. Deep learning approaches do not require this prior knowledge since they generate their own internal representations based on the training data.

\subsection{Deep learning approaches} \label{datadrivenSection}
The main advantage of deep learning approaches over the previous modeling approaches is their great modeling capacity and considerable flexibility in designing architectures. Conventional methods such as HMMs, Gaussian Processes, restricted Boltzmann machine, and dynamic random forest have been outperformed by artificial neural networks (ANNs) in modeling and generating human movements \cite{Liu2017survey, Kulsoom2022}. The underlying similarity of motion forecasting and sequence-to-sequence prediction tasks has led research in this domain toward encoder-decoder architectures. Previous research has employed Recurrent Neural Networks (RNNs) as encoders and decoders, networks that have become the standard in sequential human motion analysis \cite{Rudenko2020}. Additionally, skeletal representations are being integrated to include spatial correlations among joints for predicting full-body human movements. Shu et al. \cite{Shu2022} applied a co-attention mechanism for motion prediction that learned body joint spatial coherence and temporal evolution. Mao et al. \cite{Mao2021} proposed three autoencoder (AE) with attention, which processed motion on distinct levels: the whole body, body parts, and single joints. Cai et al. \cite{Cai2020} utilized the Discrete Cosine Transform to transform the motion into the frequency domain. The frequency components were then processed using a transformer-based architecture (global attention mechanism) in order to capture spatio-temporal correlations of the human pose. 

Variational autoencoders (VAEs) have also been applied for stochastic motion prediction, where they predict multiple and diverse motion sequences in the future from a single input sequence. Aliakbarian et al. \cite{Aliakbarian2021} accomplished this through conditional VAE, whereas Mao et al. \cite{Mao2021b} used a VAE to generate the motion of various body parts in a sequential manner. In order to enhance advances in probabilistic time series forecasting, Chung et al. \cite{Chung2015} and Fraccaro et al. \cite{Fraccaro2016} used  Recurrent Neural Networks (RNNs) to build connections between SSMs and VAEs. This led to deep SSMs, in which RNNs are used to parameterize the non-linear observation and transition models. For probabilistic forecasting, Salinas et al. \cite{Salinas2017} proposed the DeepAR, which uses auto-regressive RNNs with mean and standard deviation as output. Liu et al. \cite{Liu2020} introduced a deep SSM based on Convolutional Neural Networks that provided a unified formulation for multiple human motion systems and enabled the accurate prediction of 3D human postures.

Even though there has been a lot of progress in modeling human motions recently, with ANNs that make impressive predictions and simulations of human motions, as these approaches get more complicated, they become harder to understand and interpret their results. They can learn highly complex non-linear relationships from large datasets and surpass humans and other methods at many tasks. Nevertheless, their obscurity restricts their applicability and inspires little confidence among scientists and analysts who, for example, undertake the prognosis of movement disorders. So, while complex networks can handle activity recognition and event detection problems that put predictive accuracy above interpretability, models that can be intuitively interpreted, like analytical models, are better for applications that help people learn and improve their skills in handicrafts, industry, or sports, as well as for medical diagnostic and prognostic tools. Explainable AI (XAI) is an emerging field that focuses on addressing these challenges in deep learning through techniques such as Class Activation Mapping (CAM)\cite{Zhou2016}, Gradient-weighted Class Activation Mapping (Grad-CAM)\cite{Selvaraju2017}, Layer-wise Relevance Propagation (LRP)\cite{Bach2015}, and Testing with Concept Activation Vectors (TCAV)\cite{kim18d}. However, they have primarily been utilized for image recognition and natural language processing. This paper presents a unique work for creating interpretable models for human movement analysis using deep learning.

\section{Methodology}\label{sec:method}
This paper hypothesizes that human motion dynamics can be modeled by analytical models that can be interpreted and whose assumptions take into account the stochasticity of human motion and physical body structure. Consequently, given the nature of the hypothesis defined, the Gesture Operational Model (GOM), a hybrid stochastic-biomechanical approach based on kinematic descriptors, was proposed to model the dynamics of human movements and create interpretable representations of human motion trajectories. The authors proposed the initial concept of GOM in \cite{Manitsaris2020}, where GOM represents human movements as an SSM of a dynamic system, providing a simplified and constant mathematical formalization of the motion phenomenon. The mathematical representation of GOM permitted a more intuitive description of how body joints cooperate (spatial dynamics) and evolve over time (temporal dynamics). In this first study, GOM was demonstrated to be effective at predicting the position trajectories of the hands. Furthermore, due to the usage of a transition function, it performed well with observations obtained from varied environments and subjects without requiring extensive training datasets \cite{Olivas-Padilla2021}. This generalization capability is essential for applications requiring rapid and accurate analysis of varied human movements.

This paper presents three novel approaches for modeling human movements using full-body motion representations of GOM. In addition, an enhanced mathematical representation of GOM is used, in which parameters are no longer constant but instead time-varying. The first method calculates motion parameters using one-shot learning modeling with Kalman Filters (KF-RGOM), while the second and third applies deep learning with autoencoders (VAE-RGOM and ATT-RGOM). Each approach trains time-varying motion representations that can be utilized to explain how the inter-joint coordination changes throughout the performance of the whole movement as well as the velocity rate.

\begin{figure*}
    \centering 
    \includegraphics[width=0.8\textwidth]{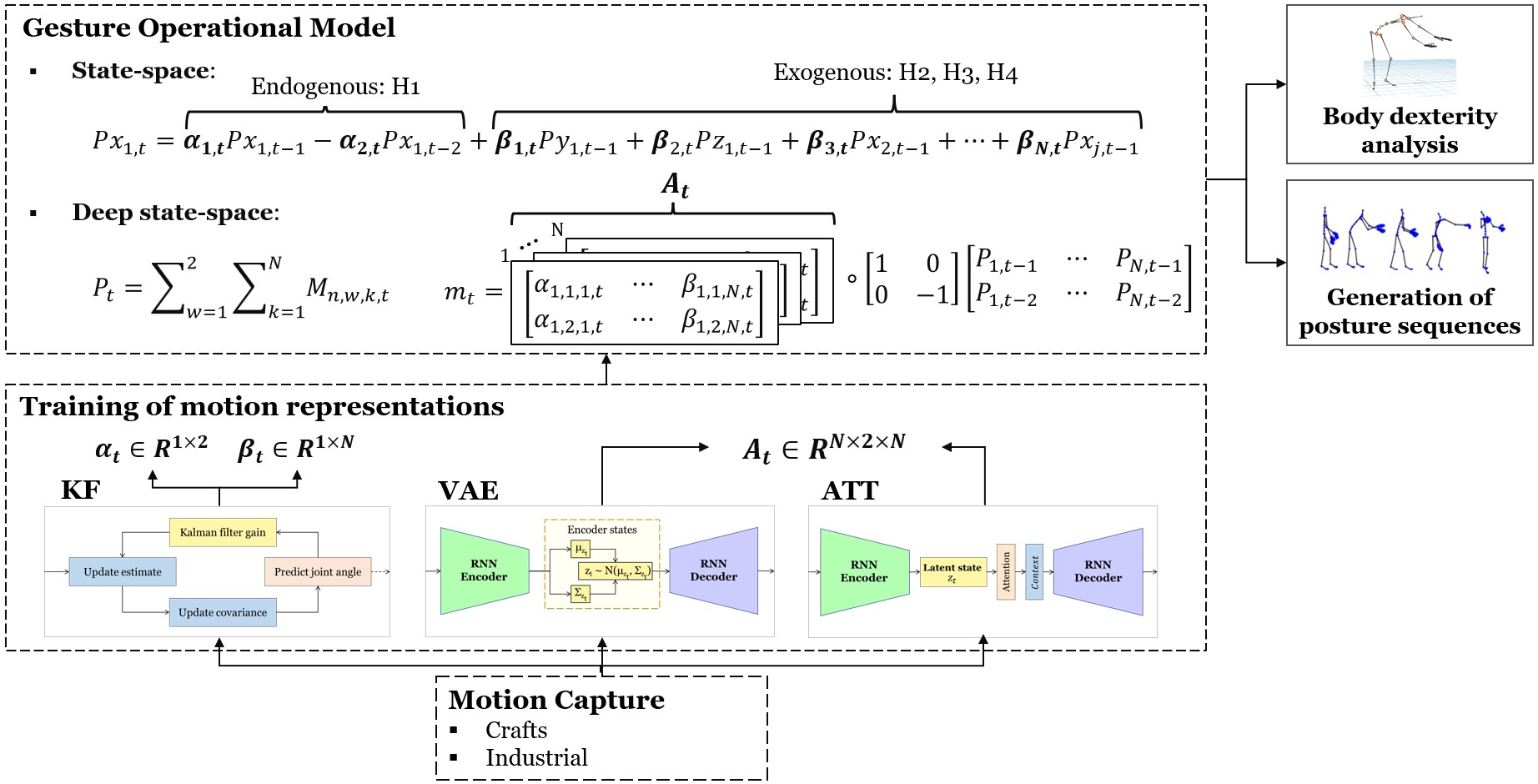} \caption{Methodology for creating explainable motion representations for body dexterity analysis and generation of human posture sequences. The motion data of industrial operators and artisans is utilized for training time-varying motion representations. Three methods are proposed for training: one-shot learning with Kalman Filters to estimate the coefficients $\alpha_t$ and $\beta_t$ of a single motion representation ($Px_{1,t}$); two methods that use deep learning with either a VAE or an Autoencoder with global attention (ATT) to automatically calculate the matrix $A_t$, which contains the coefficients of the full-body motion representations ($P_t$).}\label{fig:method}
\end{figure*}

The schematic of the methodology followed in this paper is shown in Figure \ref{fig:method}. The following Section \ref{sec:data_col} first describes the datasets used for the validation of the proposed methods. This section describes the professional movements modeled as well as the type and structure of the MoCap data that are taken into account when designing the mathematical motion representation trained by each method. Section \ref{sec:motion_rep} describes the mathematical motion representations of GOM. Then, sections \ref{sec:os_gom} and \ref{sec:dl_gom} describe the three approaches proposed for training these time-varying motion representations. Next, Section \ref{sec:bod} demonstrates how the trained motion representations provided by each method can be used for the body dexterity analysis of industrial operators and expert artisans. Finally, Section \ref{sec:sim} compares the performance in human movement generation utilizing the motion representations of each approach to that attained with the first version of GOM presented in \cite{Manitsaris2020}. 

\section{Data collection}\label{sec:data_col}
Diverse MoCap datasets have been made public for the analysis of human movements, such as HMDB51 \cite{Kuehne2011}, the MoCap dataset of Carnegie Mellon University \cite{CMU}, the KIT dataset \cite{Mandery2015}, HumanEva \cite{Sigal2010}, and MoVi \cite{Ghorbani2021}. However, most of these available datasets consist of ordinary activities and sports motions recorded inside laboratories. For that reason, seven datasets presented and further detailed in \cite{Olivas-Padilla2023} and Zenodo\footnote{Benchmark website: https://doi.org/10.5281/zenodo.5356992} were used in this paper to analyze the movements of actual operators and artisans. These datasets are composed of various professional movements captured in real-world scenarios. The recorded professional movements can be divided into two categories: industrial and crafts. Three datasets contain industrial tasks related to television assembly (TVA), packaging (TVP), and airplane assembly (APA), captured from six industrial operators. Another three correspond to motion data of four craftsmen engaged in glassblowing (GLB), silk weaving (SLW), and mastic cultivation (MSC). The last dataset corresponds to motions of different ergonomic risk levels performed by ten subjects in a laboratory (ERGD). 

For the creation of each dataset, the subjects agreed to be recorded in their actual workplace while wearing the BioMed bundle motion capture system from Nansense Inc.\footnote{Baranger Studios, Los Angeles, CA, USA}. The system is composed of a full-body suit with 52 inertial sensors strategically positioned across the torso, limbs, and hands. At a rate of 90 frames per second, the sensors measured the orientation and acceleration of body segments on the articulated spine chain, shoulders, arms, legs, and fingertips. After a recording, the Euler local joint angles on the X, Y, and Z axes were automatically calculated through the Nansense Studio's inverse kinematics.

\section{Human movement representation with state-space}\label{sec:motion_rep}
The three methods proposed are designed to parametrize mathematical motion representations based on the assumptions defined in GOM. One motion representation corresponds to an autoregressive model that learns the patterns of one motion descriptor, in the case of this paper, a local joint angle. When modeling full-body movements, there is an equation system of autoregressive models, each modeling one of the motion descriptors measured. This equation system corresponds to GOM. Suppose a human movement is depicted as a sequence of human postures ${P_t} = \left[ {{P_1},{P_2},...,{P_T}} \right] \in  \mathbb{R}^{T \times N}$. $T$ is the length of the posture sequence and $N = J \times D$, where $J$ is the number of joints measured, and $D$ is the number of dimensions that the joint's motion descriptor is decomposed. The number of models in the equation system is equal to the number of dimensions associated with a given body joint ($D$), multiplied by the number of body joints ($J$) captured with the MoCap system. Inside these $N$ models are defined four different assumptions of variables that account for the dynamic relationship between body joints and their temporal dependencies: 
\begin{itemize}
    \item[\textcolor{assumpH1}{$\rightarrow$}] \textbf{H1} Transitioning: Velocity of the movement.
    \item[\textcolor{assumpH2}{$\leftrightarrow$}] \textbf{H2} Intra-joint association: Movement of the body joint across the 3D space.
    \item[\textcolor{assumpH3}{$\leftrightarrow$}] \textbf{H3} Inter-limb synergies: Cooperation and synchronisation between body limbs.
    \item[\textcolor{assumpH41}{$\leftrightarrow$}] \textbf{H4.1} Serial \textcolor{assumpH42}{$\leftrightarrow$} \textbf{H4.2} Non-serial intra-limb mediations: Cooperation between serially and nonserially linked body joints.
\end{itemize}
Each assumption consists of a specific set of variables (local joint angles) that are parametrized and depict a particular relationship between body joints or a temporal dependency. By examining the generated coefficients and statistical significance of each variable, it can be gleaned how relevant these are according to the movement modeled and the predicted joint angles. 

\begin{figure*}
    \centering 
    \includegraphics[width=0.8\textwidth]{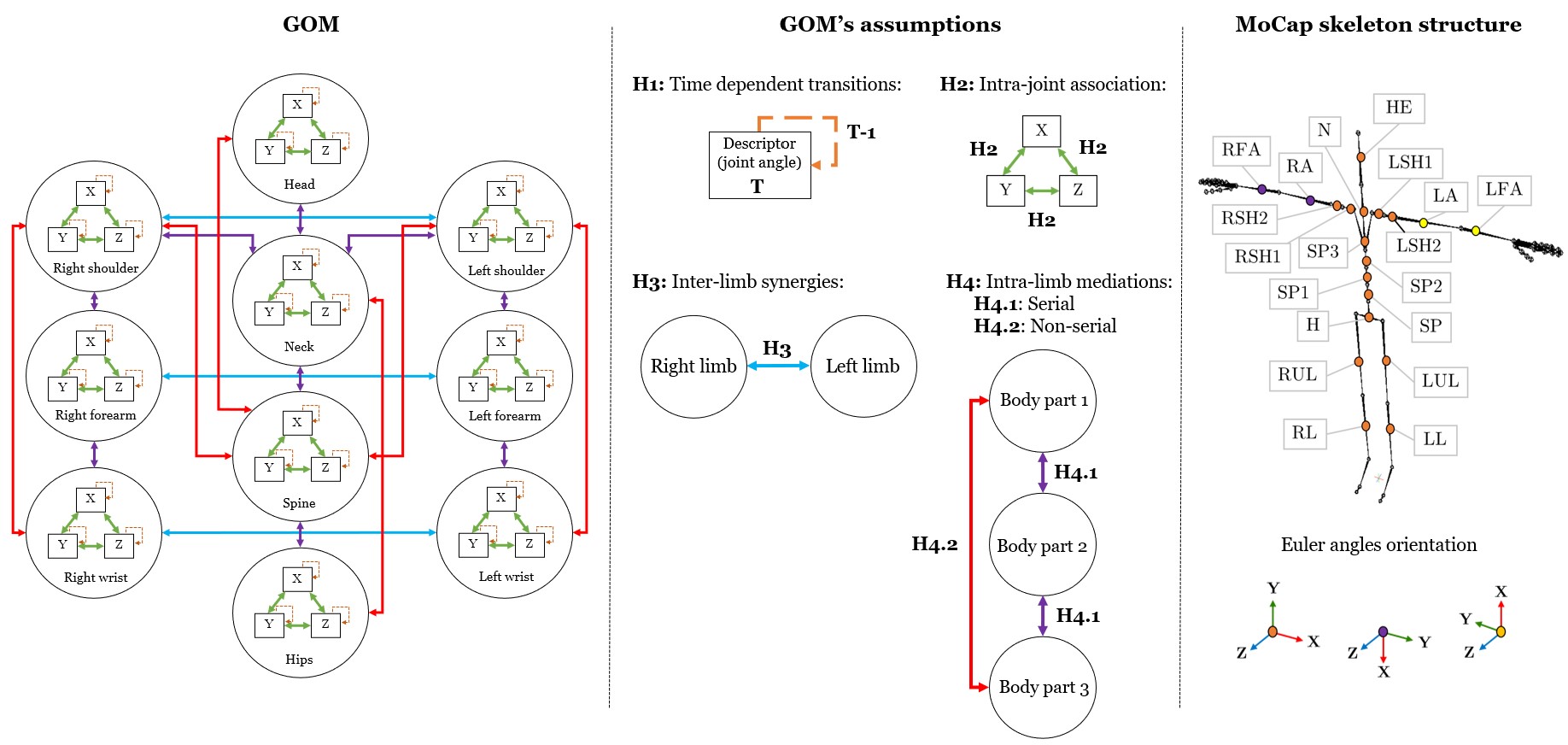} \caption{The Gesture Operational Model and assumptions. The mathematical representation of GOM is utilized to model the patterns of every joint angle of the MoCap skeleton, which assumptions are illustrated with colored arrows. }\label{fig:gomDA}
\end{figure*}

Human postures are expressed as 3D Euler joint angles in order to generate motions with subjects of various morphologies. Unlike joint positions, Euler joint angles are unaffected by identity-specific body shape. Moreover, Euler angles can be intuitively interpreted in the analytical model and clearly illustrate how human movements are conducted. A diagram of GOM and its assumptions is provided in Figure \ref{fig:gomDA}, along with the joints measured for the modeling of the full-body movements. For the purposes of this work, only measurements from 19 inertial sensors were used for the modeling. Discarding MoCap data from the fingers and feet to simplify the human motion representation. Thus, 57 joint angles were modeled in GOM.

State-space modeling was performed to create the mathematical representations, where a second-order model is designed for each motion descriptor that incorporates the assumptions as endogenous and exogenous data. Second order due to the correlation between lag values (auto-correlation) in the time series. For example, while modeling the Euler angle trajectory of the body joint $P_t$ on the \emph{X}-axis ($Px_t$), whose movement is decomposed on \emph{XYZ} axes ($Px_t$, $Py_t$, and $Pz_t$) and has an association with $j$ body parts. The two prior values are integrated into the transition model as shown in Equation \eqref{eq:gomeq1}, where $s_t$ corresponds to the state variable at time $t$. Then, exogenous data ($u_t$), corresponding to the variables from H2, H3, and H4,  are included in the observation model as illustrated in Equation \eqref{eq:gomeq2}.
\begin{equation}\label{eq:gomeq1}
s_t = A_t s_{t - 1} = \left[ {\begin{array}{*{20}{c}}
{{\alpha _{t,1}}}&0\\
0&{{\alpha _{t,2}}}
\end{array}} \right]\left[ {\begin{array}{*{20}{c}}
{Px_{1,t-1}}\\
{ - Px_{1,t-2}}
\end{array}} \right]
\end{equation}
\begin{equation}\label{eq:gomeq2}
\begin{aligned}[b]
    Px_{1,t}=\left[ {\begin{array}{*{20}{c}}
1&1
\end{array}} \right]s_t + B_t u_t = \\
\left[ {\begin{array}{*{20}{c}}
1&1
\end{array}} \right]s_t + {\beta _{t,1}}Py_{1,t-1} + {\beta _{t,2}}Pz_{1,t-1}+ \\{\beta_{t,3}}Px_{2,t-1}+\cdots+{\beta_{t,n}}Px_{j,t-1}
\end{aligned}
\end{equation}
By merging equations \eqref{eq:gomeq1} and \eqref{eq:gomeq2}, the state-space representation of the motion descriptor is obtained:
\begin{equation}
\resizebox{0.90\columnwidth}{!}{$
\begin{aligned}[b]
    Px_{1,t}=\underbrace{{\alpha_{t,1}}Px_{1,t-1}-{\alpha_{t,2}}Px_{1,t-2}}_\text{H1}+ \underbrace{{\beta_{t,1}}Py_{1,t-1} + {\beta_{t,2}}Pz_{1,t-1}}_\text{H2}+\\
   \underbrace{{\beta_{t,3}}Px_{2,t-1}+\cdots+{\beta_{t,n}}Px_{j,t-1}}_\text{H3\ or\ H4}
\end{aligned}
$}\label{eq:gomeq3}
\end{equation}
The motion representation is parametrized by time-varying coefficients $\alpha_{t}\in  \mathbb{R}^{1 \times 2}$ and $\beta_{t}\in  \mathbb{R}^{1 \times N}$. Overall, GOM comprises 57 models as Equation \eqref{eq:gomeq3}, each of which represents a joint motion trajectory.

The training of these motion representations allows for identifying and capturing the interdependence between the motion of various joints and their time transitions. Then, by examing the learned parameters, insights into how varied and complex full-body human motions are accomplished can be obtained. Additionally, by solving GOM's equation system, realistic human movement simulations can be generated. Following, it is described the three different approaches to train GOM representations.

\subsection{One-shot learning with Kalman Filters}\label{sec:os_gom}
\begin{figure*}
    \centering 
    \includegraphics[width=0.8\textwidth]{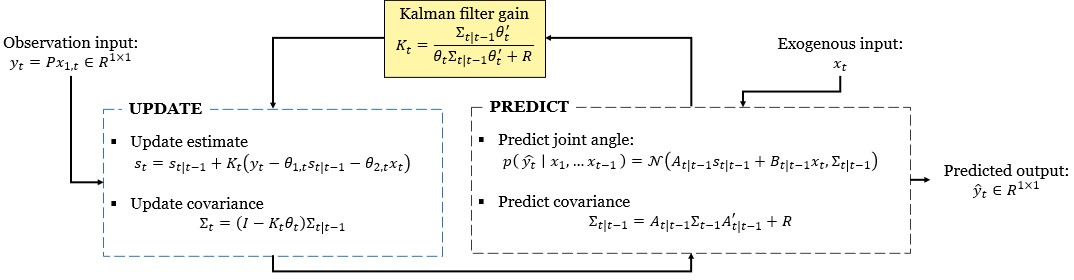} \caption{Flow chart of the iterative process of the KF while doing Maximum Likelihood Estimation (MLE) for estimating GOM's coefficients.}\label{fig:KFGOMfchart}
\end{figure*}

The first approach trains GOM representations using one-shot training with Kalman Filters (KF-RGOM). The fundamental concept is to formulate each motion representation as a separate SSM and then use KF to compute the log-likelihood of the observed motion descriptor for the given set of parameters. Suppose that the joint angle on the X-axis of a body part, $Px_{1,t}$, is modeled. Its GOM representation is the Equation \eqref{eq:gomeq3}. The observation would be the real joint angle, $y_t=Px_{1,t}$, where $Px_{1,t}\in\mathbb{R}^{1 \times 1}$, and the variables from the assumptions H2, H3, and H4 correspond to the exogenous input $x_t$. The following log-likelihood is then maximized concerning all time-varying coefficients $\theta$ and $\alpha$ and $\beta$, utilizing the KF to calculate the log-likelihood for each time $t$:
\begin{equation}\label{eq:gomkf}
\ell\left(\theta,\alpha,\beta \right)=\sum_{t=1}^{T}\log{p_\theta\left(y_1,\ldots y_{t-1}|x_1,\ldots x_{t-1}\right)}
\end{equation}
$\theta$ corresponds to the tuning parameters of the KF. The diagram of the iterative process of the Kalman filter for calculating the likelihood in \eqref{eq:gomkf} for every time $t$ is illustrated in Figure \ref{fig:KFGOMfchart}. Note that $\hat{y}_t$ corresponds to the prediction of the motion descriptor using the motion representation in \eqref{eq:gomeq3} with the estimated coefficients $\alpha$ and $\beta$.
The preceding approach is repeated for every model in GOM. This results in 57 motion representations according to the motion descriptors captured with the inertial MoCap system.

Because this approach employs one-shot training, just one movement sample per class is used to train the motion representation. This reference movement was determined using the Dynamic Time Warping (DTW) \cite{Wang2010}. This algorithm measures the similarity between two time-series. Therefore, the movement sample that was closest to all other movement samples of the same class was chosen for one-shot training. For the training of the models with KF-RGOM, it was used an Intel Core i7-8750H CPU.

\subsection{Deep learning with Autoencoders}\label{sec:dl_gom}
By taking advantage of the modeling power of ANNs, two approaches are proposed for training all motion representations of GOM simultaneously. To this end, now the observations are defined as $Y_t = P_t \in\mathbb{R}^{1 \times N}$, meaning the $N$ joint angles at time $t$ that compose the whole body posture $P_t$, and $X_t=[P_{t-1},P_{t-2}] \in\mathbb{R}^{2 \times N}$. Due to their advantages in sequence-to-sequence tasks, both frameworks use Autoencoders, where the decoders have the full-body GOM mathematical representations as the output layer. The decoder then calculates the coefficient matrix $A_t \in\mathbb{R}^{N \times 2 \times N}$:
\begin{equation}
\resizebox{0.90\columnwidth}{!}{$
    A_t=\left\{\left[\begin{matrix}\alpha_{1,1,1,t}&\cdots&\beta_{1,1,N,t}\\\alpha_{1,2,1,t}&\cdots&\beta_{1,2,N,t}\\\end{matrix}\right],\ \cdots,\left[\begin{matrix}\alpha_{N,1,1,t}&\cdots&\beta_{N,1,N,t}\\\alpha_{N,2,1,t}&\cdots&\beta_{N,2,N,t}\\\end{matrix}\right]\right\}
$}
\end{equation}
This is then utilized by the GOM equation system to produce the prediction $\hat{Y}_{t}$. Since GOM employs a second-order equation system, each element in $A_t$ corresponds to a 2D tensor with the shape $(2,N)$. $N$ because all joint angles are included in each GOM equation as an assumption (H2, H3, and H4), and two vectors as it also computed the coefficients of the transition assumptions (H1). Thus, being $X_t$ and $A_t$ tensors of shape $(2,N)$ and $(N,2,N)$, respectively, the procedure for generating $\hat{Y}_{t}$ utilizing the GOM representations in the decoder is as follows:
\begin{equation}\label{eq:dssgom1}
M_t=A_t\circ\left[\begin{matrix}1&0\\0&-1\\\end{matrix}\right] \underbrace{\left[\begin{matrix}P_{1,t-1}&\cdots&P_{N,t-1}\\P_{1,t-2}&\cdots&P_{N,t-2}\\\end{matrix}\right]}_{X_t}
\end{equation}
\begin{equation}\label{eq:dssgom2}
    \hat{Y}_{t} =\sum_{w=1}^{2}\sum_{k=1}^{N}M_{t,i,w,k}
\end{equation}
where $\circ$ is an element-wise product, $\hat{Y}_{t} \in\mathbb{R}^{1 \times N}$, and $M_t\in\mathbb{R}^{N \times 2 \times N}$. $M_t$ corresponds to GOM in a matrix form, consisting of the 57 joint angle models as Equation \ref{eq:gomeq3}.

Both Autoencoders are composed of RNNs for the encoder and decoder, representing human movements similarly to KF-RGOM, conditioning every data point at time $t$ on a hidden state at time $t-1$ as a state-space model. The observation and transition probability distributions, $p(\hat{Y}|Z,X)$ and $p(Z|X)$, are then learned maximizing the following likelihood, which approximates $\hat{Y}$ to the observed $Y$:
\begin{equation}\label{eq:dss1}
\resizebox{0.90\columnwidth}{!}{$
p_{\theta}\left(\hat{Y}_{1:T}|X_{1:T}\right)=\int_{}^{}\underbrace{p_{\theta}(\hat{Y}_{1:T}|Z_{1:T},X_{1:T})}_\text{Observation model}\underbrace{p_{\theta}(Z_{1:T}|X_{1:T})}_\text{Transition model}dZ_{1:T}
$}
\end{equation}
where $Z_{1:T}$ represents the states of the system and $X_{1:T}$ is the input that, with $Z_{1:T}$, generates the outputs $\hat{Y}_{1:T}$. In this generative model the observation model and transition model are calculated as follows:
\begin{equation}\label{eq:dss2}
    p_{\theta}(\hat{Y}_{1:T}|Z_{1:T},X_{1:T}) = \prod_{t=1}^{T}p_{\theta}(\hat{Y}_t|Z_t,X_t)
\end{equation}
\begin{equation}\label{eq:dss3}
    p_{\theta}(Z_{1:T}|X_{1:T}) = \prod_{t=1}^{T}p_{\theta}(Z_t|Z_{t-1},X_t)
\end{equation}
The parameters $\theta$ from the observation and transition models are learned during training by the decoder of each framework. However, the encoder has a different function in each approach, taking advantage of their very specific encoder-decoder architecture. In the first deep learning approach, denoted as VAE-RGOM, an architecture of a VAE is used, meaning the encoder functions as an inference network. In the second approach, designated as ATT-RGOM, the Autoencoder has incorporated a Luong attention mechanism (global) which initializes the system's state as a selected sequence of observed motion descriptors. 

\subsubsection{VAE-RGOM}
Similar to a conventional VAE, the encoder of VAE-RGOM estimates the stochastic latent states $Z$ by approximating $q_\varphi(Z_t|X_t,Y_t)$ to the true posterior distribution $p_\theta(Z_t|Z_{t-1}, X_t)$ defined by a mean $\mu_{Z_t}$ and a log covariance $\Sigma_{Z_t}$. Stochastic gradient optimization is used to train the networks, which entails first sampling $Z_t$, subsequently estimating the evidence lower bound (ELBO)\cite{kingma2013}, then the gradients for $\theta$, $\varphi$, and $A$, and lastly, updating these parameters. The loss of Equation \ref{eq:dss1} is thus equivalent to the maximum ELBO with respect to $\theta$, $\varphi$, and $A$ that results:
\begin{equation}
\resizebox{0.9\columnwidth}{!}{$
\begin{aligned}
    \ell\left(\theta,\varphi,A\right)= \sum_{t=1}^{T}{\underbrace{-\beta_{\text{VAE}}\ KL(q_\varphi(Z_t|Z_{t-1},X_t,Y_t)|\left|p_\theta\left(Z_t|Z_{t-1}, X_t\right)\right)}_\text{Regularization loss}}+ \\
\underbrace{\beta_{\text{GOM}}\mathbb{E}_{q_\varphi(Z_t|Z_{t-1},X_t,Y_t)|}\left[\log{p_\theta(Y_t|Z_t,X_t)\mathrm{\ } }\right]}_\text{Prediction loss}
\end{aligned}
$}\label{eq:vaegomloss2}
\end{equation}
In Equation \ref{eq:vaegomloss2}, KL denotes the Kullback-Leibler divergence that captures the complexity of the data; the prediction loss measures the accuracy of the model in the prediction; $\beta_{\text{VAE}}$ and $\beta_{\text{GOM}}$ correspond to tuning hyperparameters. Instead of directly sampling from $q_\varphi$ at each time step, $Z_t =\mu_{Z_t}+\epsilon \odot \Sigma_{z_t}$ is re-parametrized using samples from a normal random variable $\epsilon \sim \mathcal{N}(0,I)$. Consequently, the gradients relative to the parameters $\theta$, $\varphi$, and $A$ can be back-propagated through the encoder via the sampled $Z_t$. As the prediction loss, the mean absolute difference of all motion descriptors is used:
\begin{equation}\label{eq:predLoss}
\ell_{euler}=\frac{1}{J}\sum_{j=1}^{J}\frac{1}{D}\sum_{d=1}^{D} \left\| P_{t,j,d}-\hat{P}_{t,j,d} \right\|_{1}
\end{equation}

\begin{figure*}[]
\centering 
\sidesubfloat[]{%
  \includegraphics[width=0.45\textwidth]{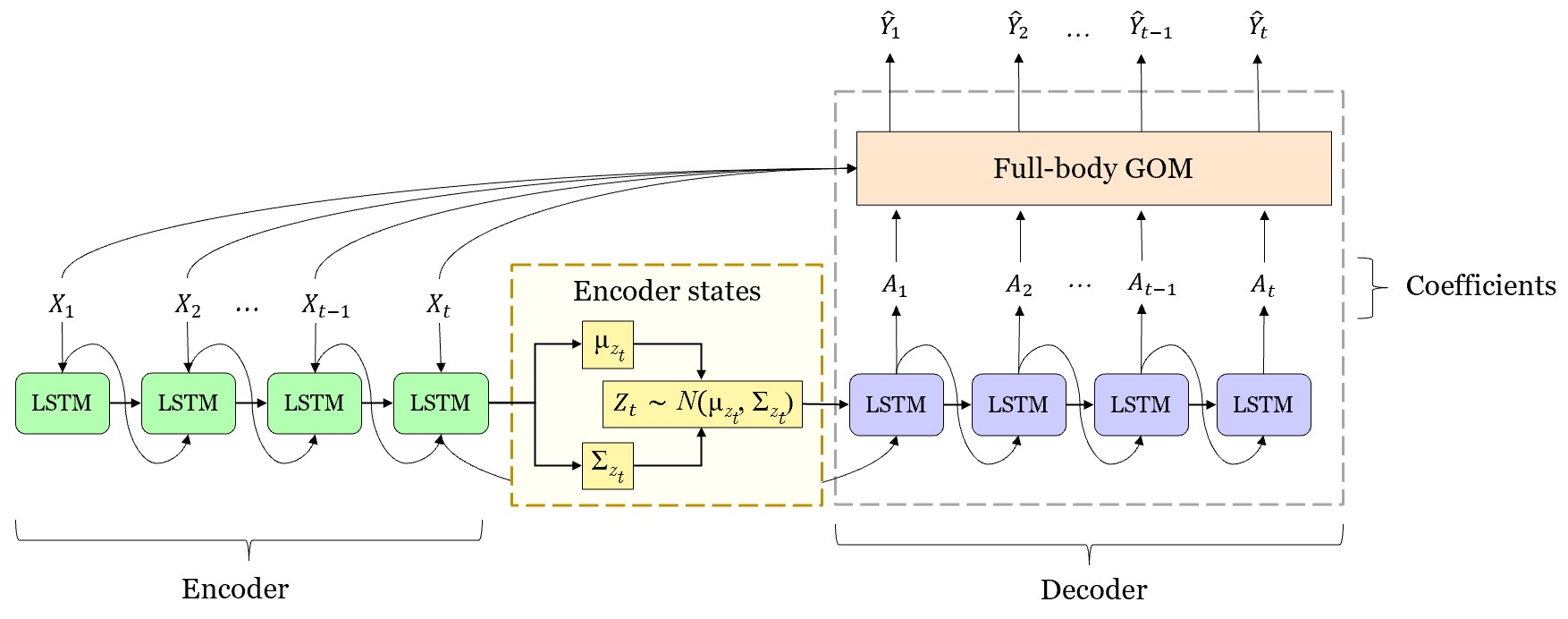}\label{fig:ae_a}
}
\sidesubfloat[]{%
  \includegraphics[width=0.45\textwidth]{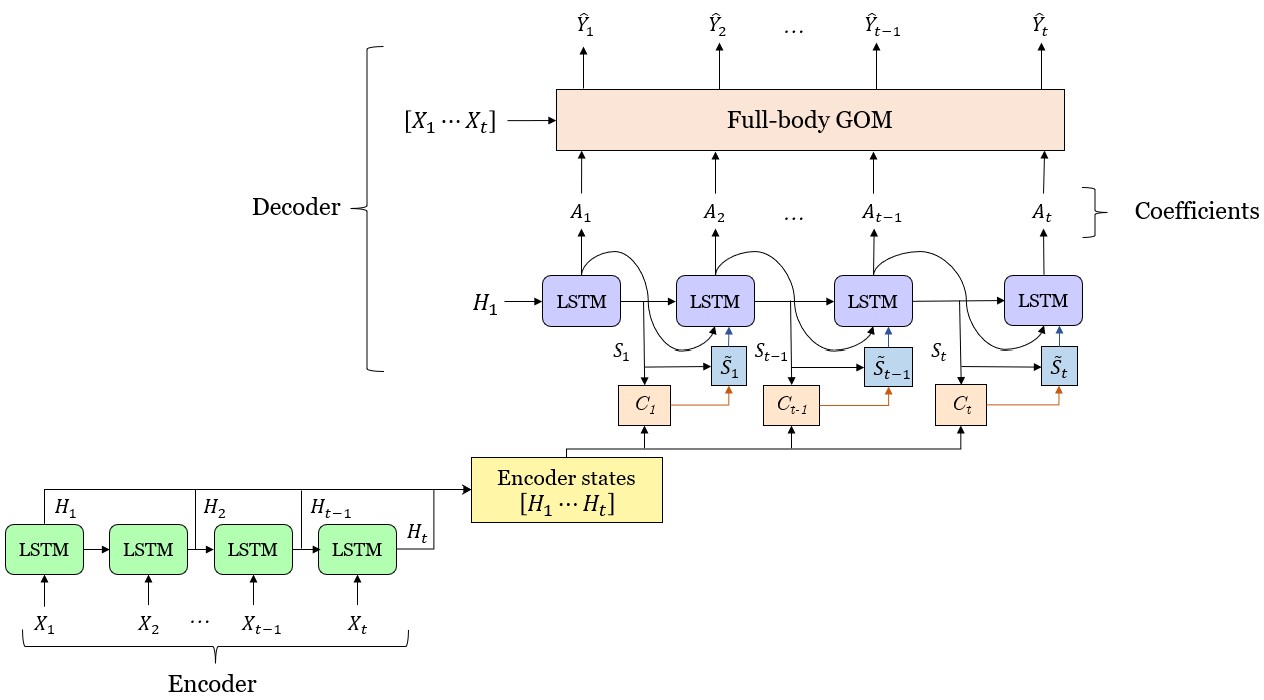}\label{fig:ae_b}
}
\caption{Schematic of each Autoencoder for estimating GOM's coefficients: \protect\subref{fig:ti_a} VAE-RGOM; \protect\subref{fig:ti_b} ATT-RGOM.}\label{fig:ae_diag}
\end{figure*}

Figure \ref{fig:ae_a} depicts a conceptual diagram of VAE-RGOM, where Long Short-Term Memory networks (LSTM) are used for both the encoder and decoder. In order to tune the network's hyperparameters, a Bayesian optimization was carried out based on the loss achieved on a validation set \cite{Snoek2012}. The best architecture is detailed in Table \ref{tab:tabvaegom}. The optimized hyperparameters included the number of units of the LSTM decoder and LSTM encoder, their learning rate, activation function, dropout rate, and recurrent dropout rate. The training was performed using backpropagation with the Adam optimizer, the initial learning rate of the optimizer was set to $1\times10^{-3}$, $\beta$ to 0.99, and the Adam parameters were $b_1 = 0.90$ and $b_2 = 0.99$. Finally, VAE-RGOM was trained and validated using all seven datasets with 5-fold cross-validation to prevent overfitting and using an NVIDIA GPU RTX 2060.

\floatsetup[figure]{style=plain,subcapbesideposition=top}

\begin{table}[t]
  \centering
    \caption{VAE-RGOM architecture.}\label{tab:tabvaegom}
\resizebox{\linewidth}{!}{%
    \begin{tabular}{lllllll} 
\hline
\multicolumn{1}{c}{\textbf{Layer}} & \multicolumn{1}{c}{\textbf{Type}} & \multicolumn{1}{c}{\begin{tabular}[c]{@{}c@{}}\textbf{Output }\\\textbf{shape}\end{tabular}} & \multicolumn{1}{c}{\textbf{Activation}} & \multicolumn{1}{c}{\textbf{Dropout}} & \multicolumn{1}{c}{\textbf{Rdrop}} & \multicolumn{1}{c}{\begin{tabular}[c]{@{}c@{}}\textbf{Input}\\\textbf{~layer}\end{tabular}}  \\ 
\hline
1                                  & Input                             & (2,57)                                                                                       & -                                       & -                                    & -                                  & -                                                                                            \\ 
\hline
2                                  & LSTM                              & (2,32)                                                                                       & Softsign                                & 0.2                                  & 0.2                                & 1                                                                                            \\ 
\hline
3 ($\mu_{Z_t}$)                      & FC                                & 2                                                                                            & Linear                                  & -                                    & -                                  & 2                                                                                            \\ 
\hline
4 ($\Sigma_{Z_t}$)                   & FC                                & 2                                                                                            & Linear                                  & -                                    & -                                  & 2                                                                                            \\ 
\hline
5 ($Z_t$)                          & Sampling                          & 2                                                                                            & -                                       & -                                    & -                                  & 3,4                                                                                          \\ 
\hline
6                                  & LSTM                              & (2,32)                                                                                       & Softsign                                & 0.2                                  & 0.2                                & 5                                                                                            \\ 
\hline
7                                  & Dropout                           & (2,32)                                                                                       & -                                       & 0.8                                  & -                                  & 6                                                                                            \\ 
\hline
8                                  & TFC~                              & (2,3249)                                                                                     & Linear                                  & -                                    & -                                  & 7                                                                                            \\ 
\hline
9 ($A_t$)                     & Reshape                           & (57,2,57)                                                                                    & -                                       & -                                    & -                                  & 8                                                                                            \\ 
\hline
10 (GOM)                           & Lambda                            & (1,57)                                                                                       & -                                       & -                                    & -                                  & 1,9                                                                                          \\
\hline
\end{tabular}}
\end{table}

\subsubsection{ATT-RGOM}
In this approach, the state-space system is parametrized by using the LSTM encoder to initialize the system's state $Z_t$ as a context vector $C_t$ of the observed joint angles. The context vector is determined by the sequence of hidden states $H_{1:T}$ to which the encoder maps the input sequence $X_{1:T}$ of length $T$. The LSTM decoder then models the state transition $p_\theta(Z_t|Z_{t-1},C_t)$ in order to update the system's state and generate the future posture $p_\theta(Y_t|Z_t,X_t)$. The Luong attention mechanism is integrated in order to capture state dynamics. The attention mechanism takes previous postures into account and maps them to attention weights ($W_t$), computed using a dot product alignment. The attention weights determine the degree to which previous hidden states $H_{1:T}$ influence future state transitions. This influence is indicated in the context vector, which is the weighted sum of the $H_i$:
\begin{equation}
C_t=\sum_{i=1}^{T}W_{t,i}H_i
\end{equation}
In accordance with the model structure of an Autoencoder with Luong attention mechanism \cite{Luong2015}, the context vector is first used to compute the attentional hidden state $\widetilde{S_t}$. The decoder then uses this state to generate the GOM's coefficients, followed by $\hat{Y}_{t}$. Figure \ref{fig:ae_b} illustrates a diagram of the ATT-RGOM structure. The network is trained to maximize the log-likelihood in Equation \eqref{eq:dss1}, considering only a prediction loss, which as VAE-RGOM, is the mean absolute difference of all motion descriptors (Equation \eqref{eq:predLoss}).

A Bayesian optimization was applied for tuning ATT-RGOM's hyperparameters as done with VAE-RGOM. Table \ref{tab:tabattgom} provides specifics on the optimized architecture. The Adam optimizer was used for the training, where the initial learning rate was set to $5\times10^{-3}$, and the Adam parameters were  $b_1 = 0.90$ and $b_2 = 0.99$. Similarly to VAE-RGOM, a 5-fold cross-validation was performed using the seven datasets and an NVIDIA GPU RTX 2060.
\begin{table}[t]
      \centering
        \caption{ATT-RGOM architecture.}\label{tab:tabattgom}
        \resizebox{\linewidth}{!}{%
        \begin{tabular}{lllllll} 
\hline
\multicolumn{1}{c}{\textbf{Layer}} & \multicolumn{1}{c}{\textbf{Type}} & \multicolumn{1}{c}{\textbf{Output shape}} & \multicolumn{1}{c}{\textbf{Activation}} & \multicolumn{1}{c}{\textbf{Dropout}} & \multicolumn{1}{c}{\textbf{Rdrop}} & \multicolumn{1}{c}{\begin{tabular}[c]{@{}c@{}}\textbf{Input}\\\textbf{~layer}\end{tabular}}  \\ 
\hline
1                                  & Input                             & (2,57)                                    & -                                       & -                                    & -                                  & -                                                                                            \\ 
\hline
\multirow{3}{*}{2}                 & \multirow{3}{*}{LSTM}             & 2.1 Output state:(2,32)                   & \multirow{3}{*}{Softsign}               & \multirow{3}{*}{0.2}                 & \multirow{3}{*}{0.2}               & \multirow{3}{*}{1}                                                                           \\
                                   &                                   & 2.2 Hidden state: 32                      &                                         &                                      &                                    &                                                                                              \\
                                   &                                   & 2.3 Cell state: 32                        &                                         &                                      &                                    &                                                                                              \\ 
\hline
3~                                 & BatchN                            & 32                                        & -                                       & -                                    & -                                  & 2.2                                                                                          \\ 
\hline
4                                  & BatchN                            & 32                                        & -                                       & -                                    & -                                  & 2.3                                                                                          \\ 
\hline
5                                  & LSTM                              & (2,32)                                    & Softsign                                & 0.2                                  & 0.2                                & 3, 4                                                                                         \\ 
\hline
6                                  & Dot                               & (2,2)                                     & -                                       & -                                    & -                                  & 2.1, 5                                                                                       \\ 
\hline
7 ($W_t$)                          & Softmax                           & (2,2)                                     & -                                       & -                                    & -                                  & 6                                                                                            \\ 
\hline
8 ($C_t$)                          & Dot                               & (2,32)                                    & -                                       & -                                    & -                                  & 2.1, 7                                                                                       \\ 
\hline
9                                  & Concatenate                       & (2,64)                                    & -                                       & -                                    & -                                  & 5, 8                                                                                         \\ 
\hline
10                                 & TFC                               & (2,3249)                                  & Linear                                  & -                                    & -                                  & 9                                                                                            \\ 
\hline
11 ($A_t$)                    & Reshape                           & (57,2,57)                                 & -                                       & -                                    & -                                  & 10                                                                                           \\ 
\hline
12 (GOM)                           & Lambda                            & (1,57)                                    & -                                       & -                                    & -                                  & 1, 11                                                                                        \\
\hline
\end{tabular}}
\end{table} 

\section{Body dexterity analysis of expert professionals} \label{sec:bod}
This section presents the analysis done over the trained GOM representations regarding their capability to explain inter-joint coordination through their mathematical assumptions. Analytical models such as these can be utilized to understand better the neurophysiological mechanisms underlying dexterity and motor learning based on the observed joint movement. Dexterity can be defined as the skill to perform a given movement or task using the hands or other body parts. The notion is to use the trained models to observe and quantify the manifestation of skill in industrial operators and expert artisans, as the learned parameters can give information about how a person moves to achieve a specific goal, such as assembling a television or making a specific piece of glass.

\subsection{Analysis of human motion representations}\label{sec:bod1}
The following are three examples of models with time-varying coefficients, each trained using one of the approaches presented in this paper. The objective is to demonstrate their capability to capture the patterns formed by each professional movement. In addition, a statistical analysis was performed on the coefficients of the motion representations to determine the significance of the motion descriptors. The null hypothesis for each coefficient is that it is equal to zero, meaning that the corresponding variable has no effect on the dependent variable. The alternative hypothesis is that the coefficient is not equal to zero, suggesting that the corresponding variable significantly affects the dependent variable. A t-test is conducted to determine the significance of the estimated coefficient in relation to its standard error. Once the p-values for each coefficient have been calculated, if the p-value is less than 0.05, the null hypothesis is rejected, concluding that the corresponding variable has a significant effect on the dependent variable. On the contrary, if the p-value is greater than 0.05, it is concluded that there is no significant evidence that the variable has an effect on the dependent variable. 

\floatsetup[figure]{style=plain,subcapbesideposition=top}
\begin{figure*}[]
\centering 
\sidesubfloat[]{%
  \includegraphics[width=0.2\textwidth]{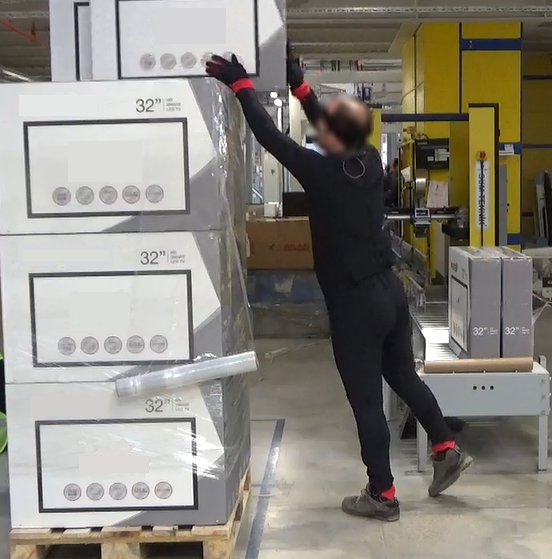}\label{fig:pm_a}
}
\sidesubfloat[]{%
  \includegraphics[width=0.2\textwidth]{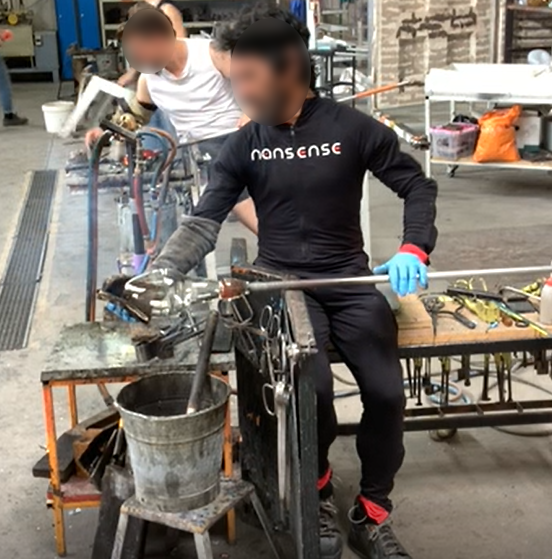}\label{fig:pm_b}
}
\sidesubfloat[]{%
  \includegraphics[width=0.2\textwidth]{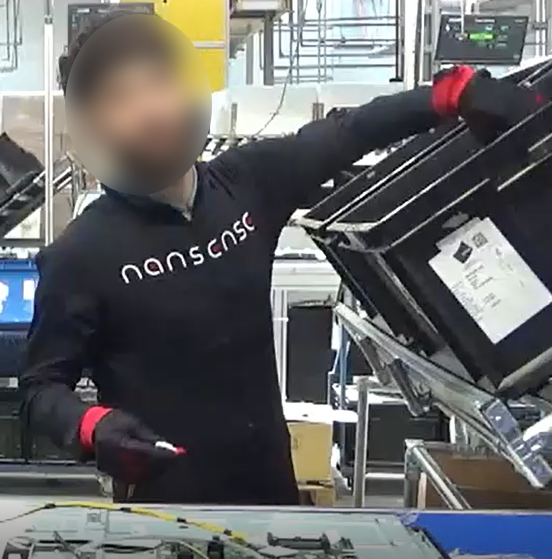}\label{fig:pm_c}
}
\caption{Visualization of the professional movements modeled and analyzed in Subsection \ref{sec:bod1}: \protect\subref{fig:pm_a} TVP: Packaging of televisions; \protect\subref{fig:pm_b} GLB: Shapping of mealting glass; \protect\subref{fig:pm_c} TVA: Grabbing circuit boards from a container during the assembly of a television.}\label{fig:pm_img}
\end{figure*}

\begin{figure*}[]
\centering 
\sidesubfloat[]{%
  \includegraphics[width=0.95\textwidth]{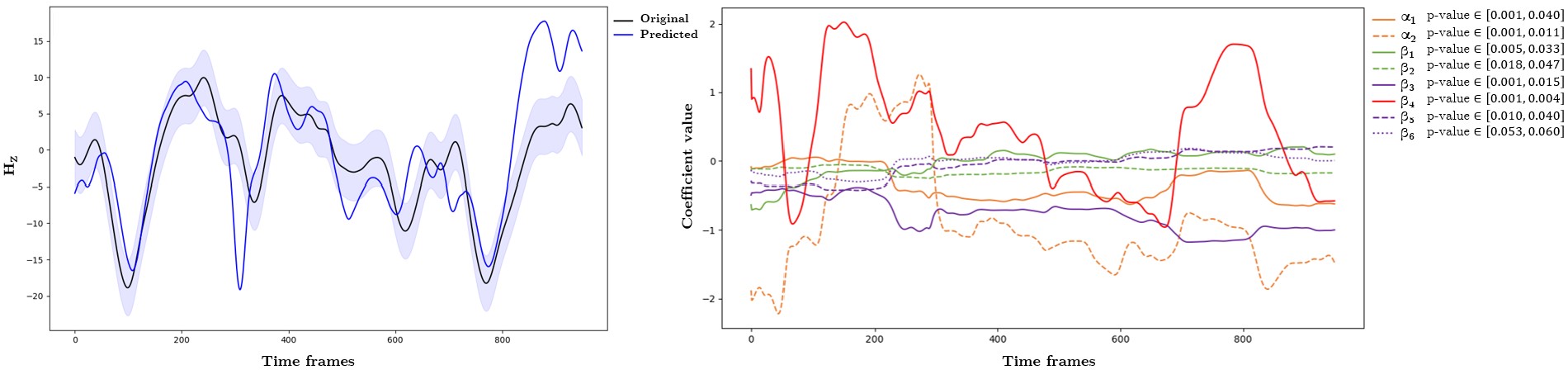}\label{fig:timevarModelDA1_a}
}\\
\sidesubfloat[]{%
  \includegraphics[width=0.95\textwidth]{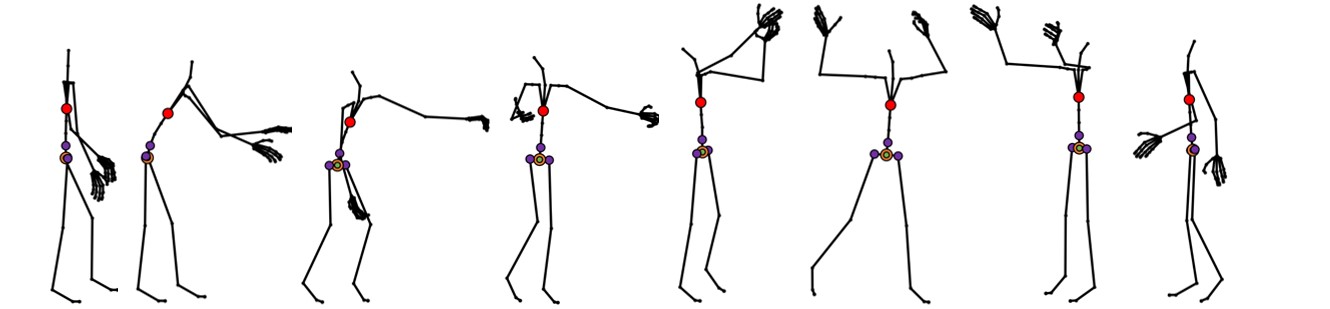}\label{fig:timevarModelDA1_b}
}
\caption{Generation of angle trajectory of $Hz$ for the packaging movement $\text{TVP}_8$: \protect\subref{fig:timevarModelDA1_a} shows the predicted angles using Equation \ref{eq:expda1}, which computed time-varying coefficients are visualized on the second plot; \protect\subref{fig:timevarModelDA1_b} illustrates the posture sequence with color annotations of the angles included as assumptions.}\label{fig:timevarModelDA1}
\end{figure*}

The first time-varying model represents a movement performed in a production line for packaging boxes containing televisions. This professional task is segmented by diverse movements, detailed in \cite{Olivas-Padilla2023} and denoted as TVP. The movement modeled consisted of an operator grabbing a box from a conveyor and placing it on the fourth level of a pallet, as illustrated in Figure \ref{fig:pm_a}. The joint angle on the Z-axis of the hips ($Hz$) is modeled and represented in Equation \eqref{eq:expda1}, which time-varying coefficients and predicted angle trajectory are illustrated in Figure \ref{fig:timevarModelDA1_a}. Additionally, Figure \ref{fig:timevarModelDA1_b} illustrates the posture sequence and highlights the angles corresponding to the equation's assumptions.
\begin{equation}
\begin{aligned}[b]
    Hz_{t}={\alpha_{1,t}}\textcolor{assumpH1}{Hz_{t-1}}+{\alpha_{2,t}}\textcolor{assumpH1}{Hz_{t-2}} \\
    +{\beta_{1,t}}\textcolor{assumpH2}{Hx_{t-1}} + {\beta_{2,t}}\textcolor{assumpH2}{Hy_{t-1}}\\
    + {\beta_{3,t}}\textcolor{assumpH41}{SPz_{t-1}}+{\beta_{4,t}}\textcolor{assumpH42}{SP3z_{t-1}}\\
    +\cdots+{\beta_{5,t}}\textcolor{assumpH41}{RULz_{t-1}}+{\beta_{6,t}}\textcolor{assumpH41}{LULz_{t-1}}
\end{aligned}\label{eq:expda1}
\end{equation}
\floatsetup[figure]{style=plain,subcapbesideposition=top}
This equation illustrates how the upper body joints of the spine and legs impact the movement of the hips during this packaging movement. The coefficients plot in Figure \ref{fig:pm_a} shows that the contribution of the joint angles to the predictions fluctuates throughout the movement. In addition to the coefficient value, if the statistical significance of the coefficients is calculated (also depicted in Figure \ref{fig:pm_a}), a temporal dependency with the two-previous value of the modeled joint angle can be identified, indicating that this movement is performed at a slow speed, as further previous values of the joint angle are still relevant to the prediction. There is an intra-joint association with the axes X and Y, indicating the movement of the hips across the 3D space, which is expected given that the operator bends and rotates his torso during the task, motions captured primarily by the axes X and Y, while Z captures lateral bending. There is no inter-limb synergy as the modeled joint angle is from the hips. However, serial intra-limb mediation is detected with $SPz$ and $RULz$, but not $LULz$, as the p-values of its coefficients are greater than 0.5. This information may reveal a behavior in the operator's gait where the right leg is prioritized for modeling the motion of the upper spine. Lastly, there is an important non-serial intra-limb mediation with $SP3z$, which has the highest coefficient values and the lowest p-values. This may be due to the fact that this particular movement focuses on moving the upper spine more than the lower spine. The operator must slightly bend their upper spine joints in order to position the box on the upper level of the pallet. Therefore, this trained motion representation demonstrates how KF-RGOM can capture the movement's primary patterns and identify joints' dependencies that cannot be observed directly.

\begin{figure*}[]
\centering 
\sidesubfloat[]{%
  \includegraphics[width=0.95\textwidth]{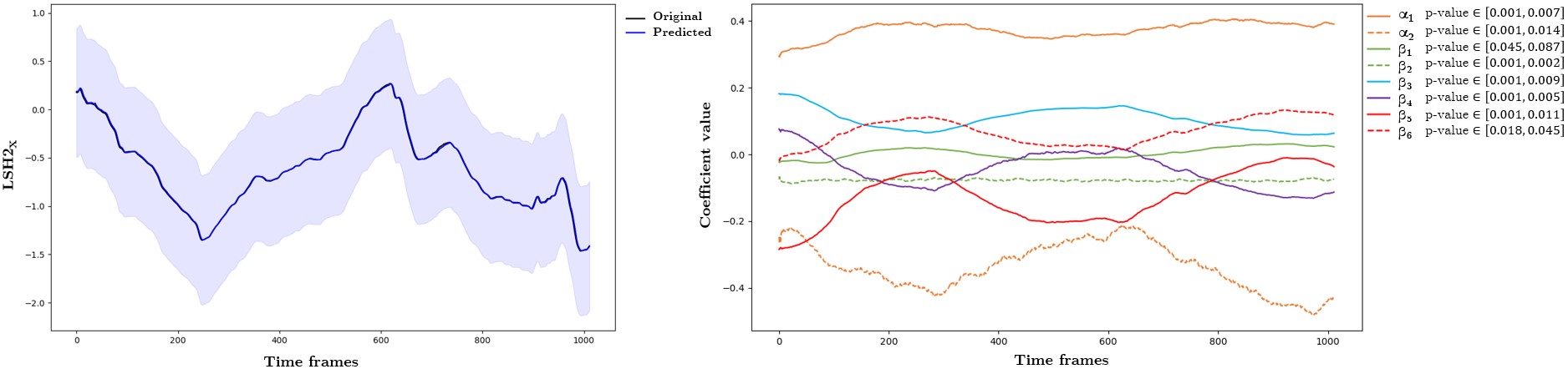}\label{fig:timevarModelDA2_a}
}\\
\sidesubfloat[]{%
  \includegraphics[width=0.95\textwidth]{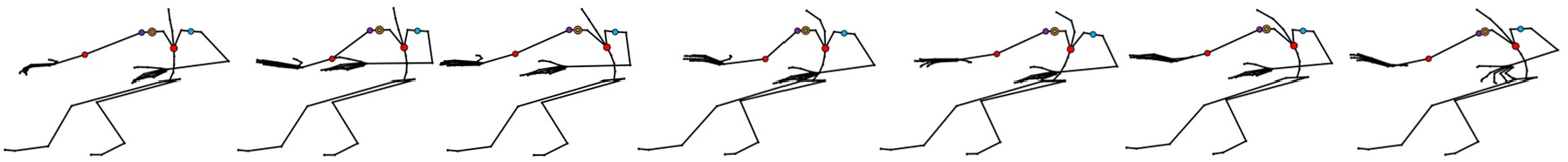}\label{fig:timevarModelDA2_b}
}
\caption{Generation of angle trajectory of $LSH2x$ for the glassblowing movement $\text{GLB}_{4}$: \protect\subref{fig:timevarModelDA2_a} shows the predicted angles using Equation \ref{eq:expda2}, which computed time-varying coefficients are visualized on the second plot; \protect\subref{fig:timevarModelDA2_b} illustrates the posture sequence with color annotations of the angles included as assumptions.}\label{fig:timevarModelDA2}
\end{figure*}

The second example is a time-varying model of $\text{GLB}_{4}$ trained using VAE-RGOM, which corresponds to the movement of shaping the glass decanter curves with a wooden block while turning the blowpipe with the right hand. The glassblower performing this movement is illustrated in Figure \ref{fig:pm_b}. The Equation \eqref{eq:expda2} represents the left shoulder's motion along the X-axis ($LSH2x_{t}$), whose time-varying coefficients and predicted angle trajectory are shown in Figure \ref{fig:timevarModelDA2_a}, and posture sequence with the assumptions in Figure \ref{fig:timevarModelDA2_b}.
\begin{equation}\label{eq:expda2}
\begin{aligned}[b]
    LSH2x_{t}={\alpha_{1,t}}\textcolor{assumpH1}{LSH2x_{t-1}}+{\alpha_{2,t}}\textcolor{assumpH1}{LSH2x_{t-2}}\\
    +{\beta_{1,t}}\textcolor{assumpH2}{LSH2y_{t-1}} + {\beta_{2,t}}\textcolor{assumpH2}{LSH2z_{t-1}}\\
    +{\beta_{3,t}}\textcolor{assumpH3}{RSH2x_{t-1}}+{\beta_{4,t}}\textcolor{assumpH41}{LAx_{t-1}}\\
    +\cdots+{\beta_{5,t}}\textcolor{assumpH42}{LFAx_{t-1}}+{\beta_{6,t}}\textcolor{assumpH42}{SP3x_{t-1}}
\end{aligned}
\end{equation}

\floatsetup[figure]{style=plain,subcapbesideposition=top}
According to the coefficient values and statistical analysis of the motion representation, a temporal dependence with both previous values of the modeled joint angle is revealed, indicating a moderate speed in the performance of the movement. There is an intra-joint association for all time series with $LSH2z$, but not in part with $LSH2y$, which is reflected in its small coefficient values and high p-values. The movement presents an inter-limb synergy with $RSH2x$. This suggests cooperation and coordination between the two arms. As visualized in figure \ref{fig:timevarModelDA2_b} and \ref{fig:pm_b}, the glassblower manipulated the molten glass with one arm while rotating it with the other. This action requires synchronization between both arms, which is captured in the trained motion representation. A non-serial intra-limb mediation with $LFAx$ is detected, which is expected given that the glassblower controls the blowpipe with his left forearm. The model shows a mediation with $SP3x$ but with p-values near the significance threshold of 0.05. Accordingly, the movement of the upper part of the spine is crucial when performing this particular movement in glassblowing. This may also be observed in Figure \ref{fig:timevarModelDA2_b}, where the glassblower moves his arms while bending his torso back and forth.

\begin{figure*}[]
\centering 
\sidesubfloat[]{%
  \includegraphics[width=0.95\textwidth]{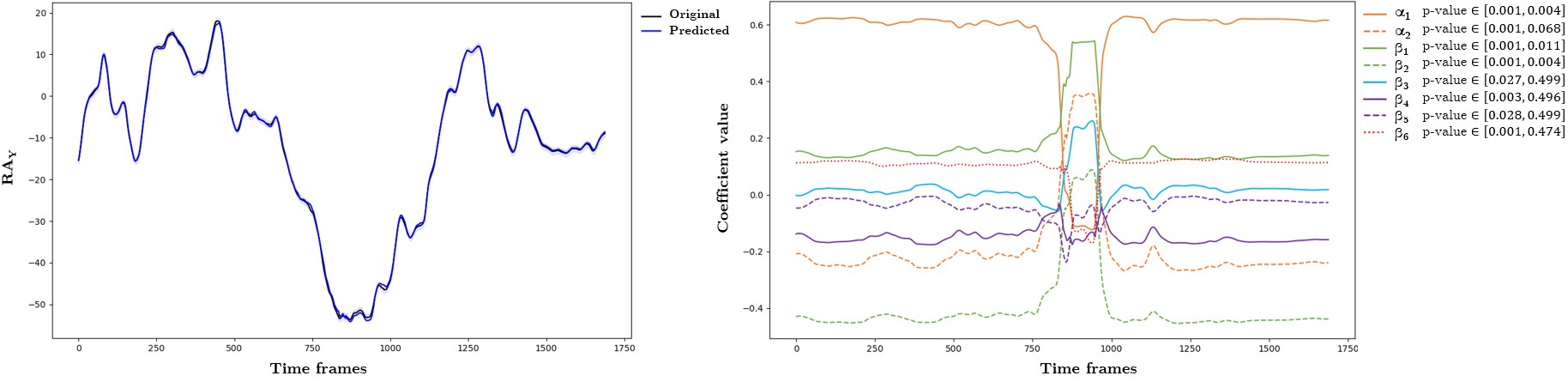}\label{fig:timevarModelDA3_a}
}\\
\sidesubfloat[]{
  \includegraphics[width=0.95\textwidth]{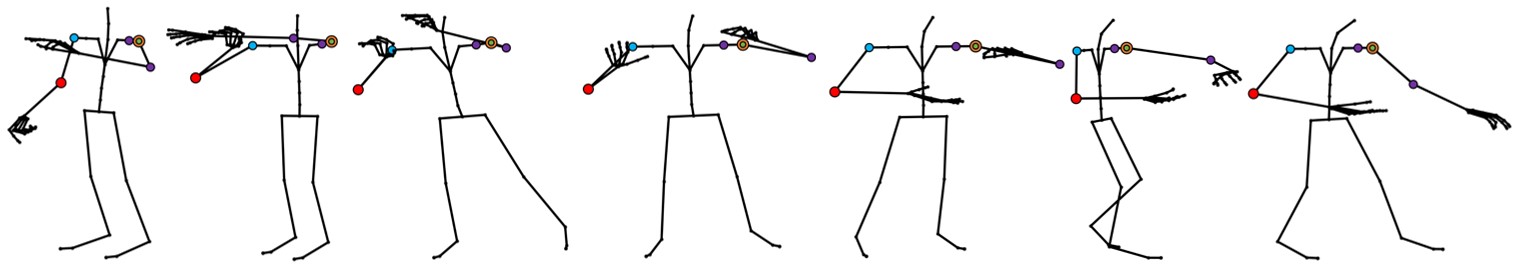}\label{fig:timevarModelDA3_b}
}
\caption{Generation of angle trajectory of $RAy$ for the assembly movement $\text{TVA}_{1}$: \protect\subref{fig:timevarModelDA3_a} shows the predicted angles using Equation \ref{eq:expda3}, which computed time-varying coefficients are visualized on the second plot; \protect\subref{fig:timevarModelDA3_b} illustrates the posture sequence with color annotations of the angles included as assumptions.}\label{fig:timevarModelDA3}
\end{figure*}

The last example is the time-varying model for the professional movement $\text{TVA}_1$ trained using ATT-RGOM. The movement consists of an operator grabbing a circuit board from a container, as seen in Figure \ref{fig:pm_c}. The joint angle on the Y-axis of the right arm ($RAy$) is modeled and represented in Equation \eqref{eq:expda3}. Figure \ref{fig:timevarModelDA3_a} shows the plots of the predicted joint angle using the trained motion representation and the equation's time-varying coefficients. The posture sequence highlighting the joint angles included as assumptions in Equation \eqref{eq:expda3} is illustrated in Figure \ref{fig:timevarModelDA3_b}.
\begin{equation}
\begin{aligned}[b]
    RAy_{t}={\alpha_{1,t}}\textcolor{assumpH1}{RAy_{t-1}}+{\alpha_{2,t}}\textcolor{assumpH1}{RAy_{t-2}} \\
    +{\beta_{1,t}}\textcolor{assumpH2}{RAx_{t-1}} + {\beta_{2,t}}\textcolor{assumpH2}{RAz_{t-1}}\\
    + {\beta_{3,t}}\textcolor{assumpH3}{LAy_{t-1}}+{\beta_{4,t}}\textcolor{assumpH41}{RSH1y_{t-1}}\\
    +\cdots+{\beta_{5,t}}\textcolor{assumpH41}{RFAy_{t-1}} +{\beta_{6,t}}\textcolor{assumpH42}{LFAy_{t-1}}
\end{aligned}\label{eq:expda3}
\end{equation}
Figure \ref{fig:timevarModelDA3} depicts the predicted joint angle trajectory using Equation \ref{eq:expda3}, its computed time-varying coefficients, and the posture sequences with highlighted joint angles incorporated as assumptions in the equation.
\floatsetup[figure]{style=plain,subcapbesideposition=top}
In this example, the time-varying coefficients and p-values indicate a time dependence, with both previous values implying a moderate-speed motion; however, at certain periods of the time series, the values two time steps prior to the prediction were not significant. The model exhibits an intra-joint association with the X and Z axes of RA ($RAx$ and $RAz$), meaning a movement of the right arm in the 3D space. There is also an inter-limb synergy, but based on the coefficients and p-values, it was not present during certain periods of the time series. This indicates that there was not always a relationship between the movements of the right arm and the left arm, which could be the case if the operator was simply holding the card with the right hand. A synergy occurred when he passed the circuit from one hand to the other, as depicted in the second-last posture of the sequence in Figure \ref{fig:timevarModelDA3_b}. Lastly, there is a serial intra-limb mediation with $RSH2y$ and $RFAy$, but it was not present throughout the entire time series. Specifically, $RSH2y$ and $RFAy$ were not significant on transitions where the operator is walking toward the container or just standing and holding the circuit board for a moment. Similarly, non-serial intra-limb mediation with $LFAy$ was only observed when the operator moved both arms simultaneously.  

The preceding examples showed how the trained analytical models can be utilized to explain the physical dexterity of industrial operators and expert artisans. The models emphasized the key motion descriptors associated with and contributing to complex whole-body movements. This information can later be utilized to test skill acquisition strategies. A novice can learn to make precise movements by minimizing the variability of their motion representations compared to those of professional artisans or operators.

\subsection{Selection of the most significant sensors to maximize recognition accuracy}
This section illustrates how the results from the body dexterity analysis can be applied to find the best motion descriptors for modeling and recognizing a set of human movements. For this paper, the best motion descriptors for recognizing the professional gestures from each of the seven datasets are found and presented. After performing the statistical analysis, the number of times a motion descriptor (assumption) is statistically significant for all equations that comprise GOM is counted. Then for the selection, different combinations of descriptors considered most frequently significant are utilized only for training in an all-shots approach. Because a single inertial sensor gives three joint angles, all of the sensor's joint angles are used for recognition if at least one is among the joint angles that are more often significant in all gestures of a dataset. 

For the recognition of the professional gestures from the seven datasets utilizing different sensor combinations, HMMs were trained. To properly train the HMMs, a gesture vocabulary containing the gestures with the most samples was specified for each dataset. The total number of motion classes for TVA, APA, and ERGD were four, three, and 28, respectively. The TVP, GLB, and MSC gesture vocabularies contained only gestures with at least seven samples. Therefore, their respective gesture vocabularies included five, seven, and six classes of gestures. Regarding SLW, the gesture vocabulary consisted of only three classes of silk weaving on a loom. 

All gestures from the vocabularies were modeled using each of the three approaches and subjected to the selection process described previously to compare the selection obtained with the models of each approach. Then, the sensor configurations with the best performance were compared to the recognition performance obtained by utilizing motion data from all sensors. The performance yielded using the selection of sensors obtained with constant GOM motion representations, proposed initially by Manitsaris et al. \cite{Manitsaris2020} and denoted as KF-GOM, was also included for comparison. In KF-GOM, constant GOM representations are learned using one-shot training with Kalman Filters. Additionally, the recognition performance using a minimal set of two sensors was also computed and compared. This minimal set consisted of two hand-picked sensors that provided the Euler joint angles of the right forearm (RFA) and hips (H). The sensor positioned on the right forearm was chosen since the majority of individuals in all datasets were right-handed, and the sensor placed on the hips because all spinal movement originates from the hips. The purpose of these comparisons is to assess the method's capability to select the subset of sensors that achieves superior recognition performance over configurations that include all 52 inertial sensors or a manually picked set of sensors.

\begin{figure*}
    \centering 
    \includegraphics[width=0.8\textwidth]{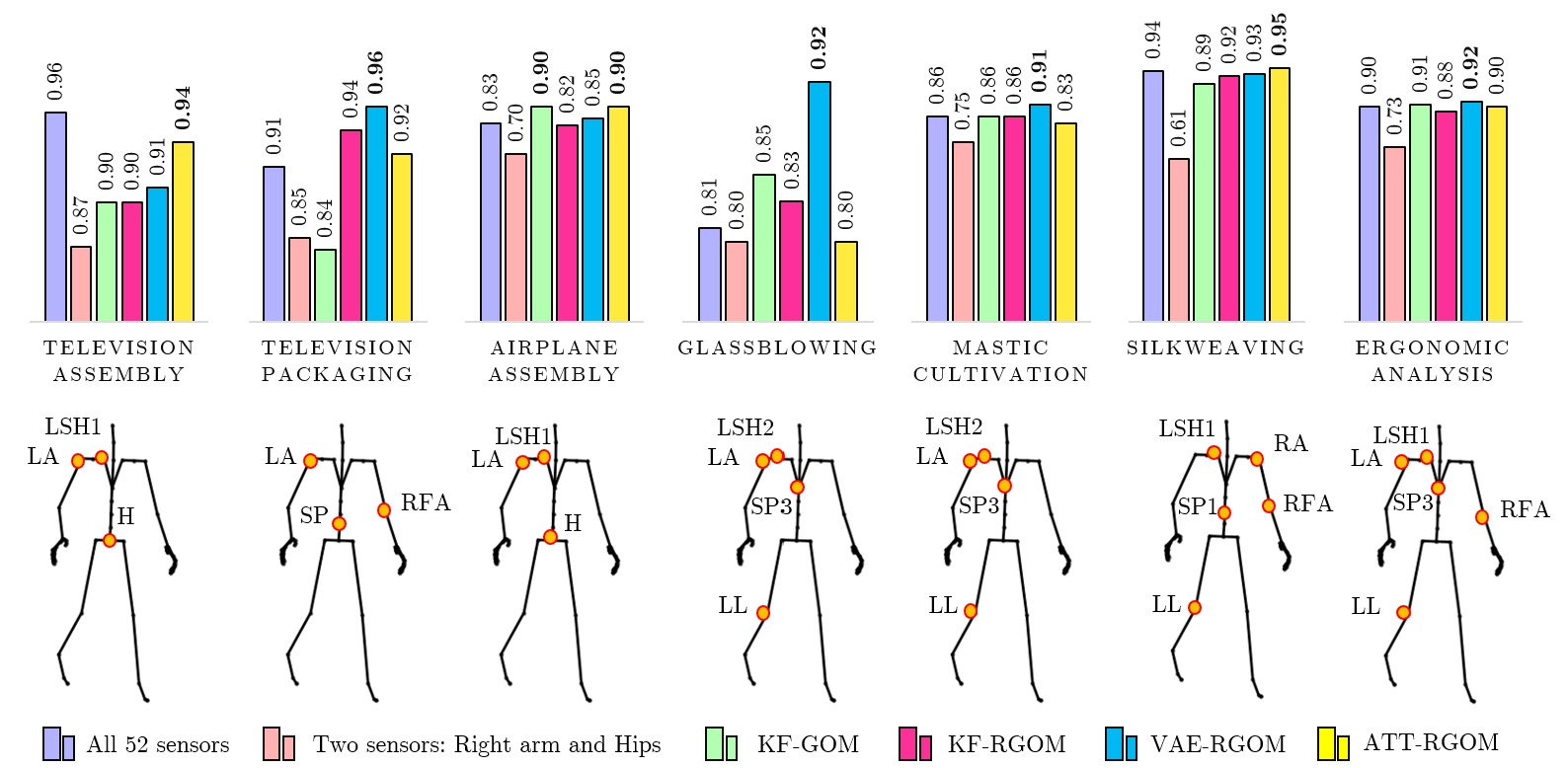} 
    \caption{F1-score achieved with each dataset and the best minimal sensor configuration.}\label{fig:plotF1}
\end{figure*}

The ergodic and left-to-right HMM  topologies and a different number of hidden states were tested to determine the best HMM settings for the gesture vocabulary defined in each dataset. Left-to-right HMM topology produced the best results for all recognition problems. Concerning the number of hidden states, it was defined for the HMMs of TVA and ERGD with seven states, TVP with six states, APA, GLB, and MSW with eight states, and SLW with three states. Figure \ref{fig:plotF1} displays the calculated F1-scores for each recognition problem and the best sensor configuration identified.

The relevance of the sensors selected based on each approach for each dataset was demonstrated by the superior or similar recognition performance attained compared to using all sensor data. By observing the graphs from Figure \ref{fig:plotF1}, the best minimal set for recognizing gestures from television assembly (TVA) and silk-weaving (SLW) was determined by using the representations estimated by ATT-RGOM. These sensors achieved comparable results to using all sensors' data, having the ATT-RGOM set an F1-score of 0.949 for TVA, whereas all sensors set had an F1-score of 0.966. Then, for SLW, the ATT-RGOM sensor set performed better than all other sensor sets. VAE-RGOM representations provided the best sensor set for recognizing the gestures from the packaging dataset (TVP), with an F1-score of 0.966. When comparing the selected sensors from the VAE-RGOM and ATT-RGOM sets, it is observed that the VAE-RGOM set included a sensor from the lower spine (SP), shown in the corresponding skeleton in Figure \ref{fig:plotF1}, which was better for recognition than the SP1 selected by ATT-RGOM. This sensor enhanced the ability to distinguish between the movements of placing a box on the first and second levels. The recognition of gestures from airplane assembly (APA) was better using the data provided by the sensor sets of KF-RGOM and ATT-RGOM, although the ATT-RGOM sensor set contained fewer sensors, which are illustrated in the corresponding skeleton of Figure \ref{fig:plotF1}. The set selected using ATT-RGOM representations attained an F1-score of 0.905, outperforming the set containing all sensor data. The APA gestures were the most challenging to recognize. This may be because the gestures in this vocabulary are more complex and prolonged. In addition, there is a substantial intra-class variance due to the fact that just one airplane structure was constructed for this dataset. There were no repetitions in which the pneumatic hammer was positioned in the same location more than once.

The glassblowing gestures performed in GLB were better recognized using the data from the sensor sets estimated using VAE-RGOM representations, reaching an F1-score of 0.923. According to all GOM representations, the shoulders contribute the most to the execution of glassblowing gestures, which is why the two-sensors configuration performed the worst. In addition, using the motion data measured from the left calf, which VAE-RGOM included, the recognition performance improved by at least 7\% in the F1-score compared to all other sets.

The sensors picked using VAE-RGOM representations were also the most effective at discriminating motions from mastic cultivation (MSC) and the ergonomic analysis dataset (ERGD). These were the gesture vocabularies with the greatest number of classes. Because similar postures are performed on the movements of each dataset, the selected sensors for each recognition problem are almost the same. The VAE-RGOM sensor set attained an F1-score of 0.913 for MSC. Then, for ERGD, the VAE-RGOM sensor set achieved an F1-score of 0.926. The poor performance of the two-sensor configuration for MSC and ERGD may have been caused by its inability to differentiate between movements that differ only in the posture of the legs, as the motion data from the hips was insufficient. In conclusion of this subsection, the minimal sets of motion descriptors to measure for recognizing the gestures from the seven datasets were successfully identified as similar or better performances were achieved than using all motion data of the inertial MoCap system.

\section{Generation of professional movements}\label{sec:sim}
This section evaluates KF-RGOM, VAE-RGOM, and ATT-RGOM in terms of their ability to generate realistic human movements. Their performance is compared to that of the first method, KF-GOM, which learned constant motion representations via one-shot training.

\begin{figure}
    \centering 
    \includegraphics[width=0.9\textwidth]{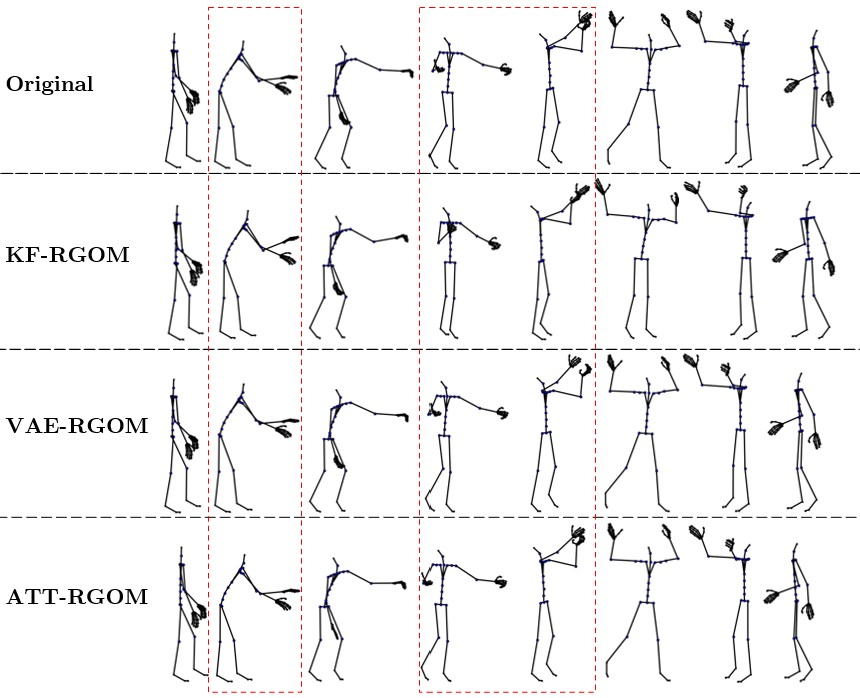} \caption{Visual comparison of generated posture from TVP and its ground-truth. The operator grabs a box from a conveyor and places it on the fourth level of a pallet.}\label{fig:postureTVP}
\end{figure}

\begin{figure}
    \centering 
    \includegraphics[width=0.9\textwidth]{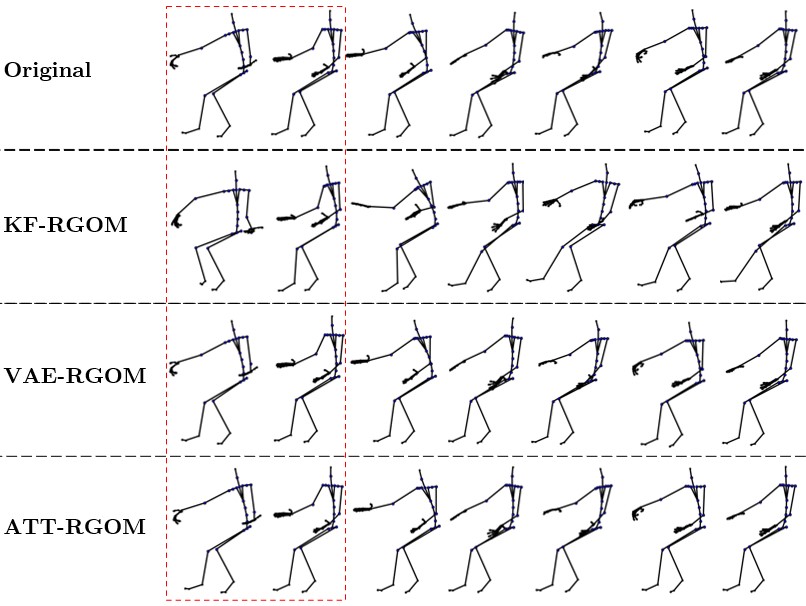} \caption{Visual comparison of generated posture from GLB and its ground-truth. The glassblower rotates the blowpipe with the left hand while shaping the glass with the right.}\label{fig:postureGLB}
\end{figure}

\begin{figure}
    \centering 
    \includegraphics[width=0.9\textwidth]{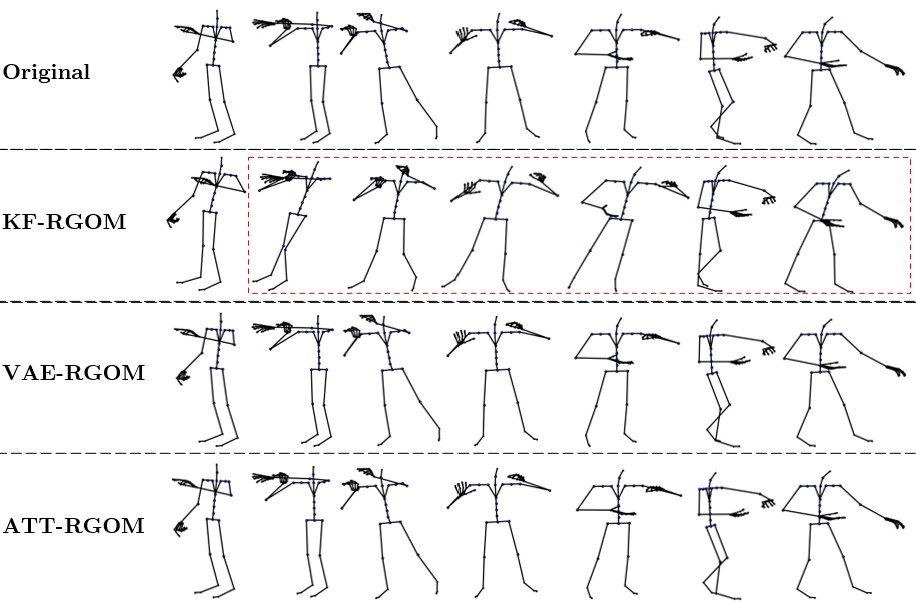} \caption{Visual comparison of generated posture from TVA and its ground-truth. The operator grabs a circuit card from a container.}\label{fig:postureTVA}
\end{figure}

Each of the four approaches generated one time step per iteration. After generating all the time frames of a movement, the artificial movement was compared with the original for evaluation. The KF-RGOM and KF-GOM were trained using a reference movement per class, whereas the Autoencoders used all movements of the datasets in a 5-fold cross-validation. A validation set was used to estimate the neural networks' hyperparameters, while the test set was utilized for evaluation. Tables \ref{tab_mae}, \ref{tab_rmse}, and \ref{tab_theil} present the average Mean Absolute Error (MAE), Root Mean Squared Error (RMSE), and Theil Inequality Coefficient ($U_1$) \cite{Leuthold1975}, respectively, achieved with each dataset and approach. All movements were generated with the respective motion representation of their class, then the MAE, RMSE, and U1 were calculated between the generated movement and the original. MAE and RMSE measure the capacity of the model to generate diverse samples without considering their quality, while $U_1$ is a bounded statistic $(0,1)$, with perfect forecasting resulting in $U_1 = 0$ (i.e., $Y_t = \hat{Y}_t$ $\forall t$). To complement the analysis of the generation of full-body movements, visual comparisons between the quality of the generated sequences and the ground truth sequence are offered in figures \ref{fig:postureTVP}, \ref{fig:postureGLB}, \ref{fig:postureTVA}. The red boxes show the biggest variations presented in the generated movements at different temporal windows.

\begin{table}[]
\centering
\caption{Average MAE for each dataset.}\label{tab_mae}
\resizebox{\linewidth}{!}{%
\begin{tabular}{lllll}
\hline
Dataset & KF-GOM          & KF-RGOM        & VAE-RGOM      & ATT-RGOM      \\ \hline
TVA     & 16.830 ($\sigma$: 15.379) & 6.938 ($\sigma$: 0.459)  & 0.093 ($\sigma$: 0.016) & 0.191 ($\sigma$: 0.024) \\
TVP     & 7.947 ($\sigma$: 5.278)   & 9.867 ($\sigma$: 5.592)  & 0.213 ($\sigma$: 0.058) & 0.398 ($\sigma$: 0.085) \\
APA     & 3.312   ($\sigma$: 2.816) & 10.946 ($\sigma$: 0.664) & 0.091 ($\sigma$: 0.019) & 0.203 ($\sigma$: 0.048) \\
GLB     & 19.211 ($\sigma$: 9.981)  & 12.916 ($\sigma$: 3.665) & 0.119 ($\sigma$: 0.044) & 0.220 ($\sigma$: 0.066) \\
SLW     & 14.267 ($\sigma$: 7.739)  & 9.207 ($\sigma$: 3.307)  & 0.115 ($\sigma$: 0.041) & 0.246 ($\sigma$: 0.078) \\
MSC     & 23.313 ($\sigma$: 14.514) & 16.002 ($\sigma$: 5.887) & 0.247 ($\sigma$: 0.101) & 0.457 ($\sigma$: 0.126) \\
ERGD    & 13.461 ($\sigma$: 8.699)  & 13.569 ($\sigma$: 4.931) & 0.095 ($\sigma$: 0.052) & 0.198 ($\sigma$: 0.085) \\ \hline
\end{tabular}}
\end{table}

\begin{table}[]
\centering
\caption{Average RMSE for each dataset.}\label{tab_rmse}
\resizebox{\linewidth}{!}{%
\begin{tabular}{lllll}
\hline
Dataset & KF-GOM          & KF-RGOM        & VAE-RGOM      & ATT-RGOM      \\ \hline
TVA     & 32.438 ($\sigma$: 27.247)   & 14.903 ($\sigma$: 1.574)    & 0.962 ($\sigma$: 0.430) & 1.126 ($\sigma$: 0.410)  \\
TVP     & 15.978 ($\sigma$: 5.614)    & 20.937 ($\sigma$: 2.391)    & 3.231 ($\sigma$: 1.402) & 3.339 ($\sigma$: 1.389) \\
APA     & 17.814 ($\sigma$: 17.652)   & 19.127 ($\sigma$: 10.266)   & 0.885 ($\sigma$: 0.147) & 1.034 ($\sigma$: 0.146) \\
GLB     & 42.918 ($\sigma$: 22.873)   & 29.097 ($\sigma$: 10.373)   & 2.049 ($\sigma$: 1.384) & 2.204 ($\sigma$: 1.388) \\
SLW      & 28.787 ($\sigma$: 13.616)  & 22.868 ($\sigma$: 10.798)   & 0.467 ($\sigma$: 0.287) & 0.721 ($\sigma$: 0.328)\\
MSC     & 51.455 ($\sigma$: 19.791)   & 36.828 ($\sigma$: 22.618)   & 3.103 ($\sigma$: 2.043) & 3.311 ($\sigma$: 1.980)\\
ERGD    & 21.732 ($\sigma$: 13.926)   & 15.126 ($\sigma$: 11.006)   & 1.134 ($\sigma$: 0.758) & 1.279 ($\sigma$: 0.782)\\ \hline
\end{tabular}}
\end{table}

\begin{table}[]
\centering
\caption{Average $U_1$ for each dataset.}\label{tab_theil}
\resizebox{\linewidth}{!}{%
\begin{tabular}{lllll}
\hline
Dataset & KF-GOM          & KF-RGOM        & VAE-RGOM      & ATT-RGOM      \\ \hline
TVA              & 0.427 ($\sigma$: 0.368) & 0.384 ($\sigma$: 0.057)  & 0.015 ($\sigma$: 0.005)   & 0.023 ($\sigma$: 0.005)    \\ 
TVP              & 0.195 ($\sigma$: 0.068) & 0.125 ($\sigma$: 0.073)  & 0.019 ($\sigma$: 0.004)   & 0.025 ($\sigma$: 0.003)    \\ 
APA              & 0.310 ($\sigma$: 0.192) & 0.210 ($\sigma$: 0.101)  & 0.009 ($\sigma$: 0.003)   & 0.016 ($\sigma$: 0.003)    \\ 
GLB              & 0.540 ($\sigma$: 0.221) & 0.292 ($\sigma$: 0.123)  & 0.026 ($\sigma$: 0.015)   & 0.028 ($\sigma$: 0.015)    \\ 
SLW              & 0.390 ($\sigma$: 0.341) & 0.201 ($\sigma$: 0.102)  & 0.043 ($\sigma$: 0.016)   & 0.048 ($\sigma$: 0.013)    \\ 
MSC              & 0.586 ($\sigma$: 0.262) & 0.361 ($\sigma$: 0.099)  & 0.024 ($\sigma$: 0.010)   & 0.030 ($\sigma$: 0.009)    \\ 
ERGD             & 0.394 ($\sigma$: 0.281) & 0.2742 ($\sigma$: 0.056) & 0.010 ($\sigma$: 0.003)   & 0.015 ($\sigma$: 0.003)    \\
\hline
\end{tabular}}
\end{table}

The experiments suggest that by solving the simultaneous equations that compose GOM, it is possible to accurately generate diverse human movements using Euler joint angles as motion descriptors. Overall, time-varying GOM representations are tolerant of slight variations in human movements and offsets between movements of the same class produced by varying recording conditions (different subjects or different recording days). Furthermore, implementing time-varying coefficients increased the modeling performance of GOM, especially for movements with greater variance and longer duration, such as those conducted during glassblowing (GLB). This is due to the fact that coefficients adapted to the change in mediations between the dependent variables and their assumptions throughout the whole time series. 

Across all seven datasets, the time-varying parameter models trained by Autoencoders performed the best. Two arguments were deduced as to why this improvement in performance. First, both approaches used motion data from all seven datasets for training, which would have allowed them to map diverse relationships between assumptions and dependent variables and accurately estimate the optimal coefficients for one-step prediction. The variability between the training and test movements may account for the errors exhibited by both KF-GOM and KF-RGOM, as only one reference movement is used for training. Accordingly, it can be inferred that the quality of the simulations by one-shot training depends on the recorded person's ability to replicate their movements while repeatedly performing the same activity. Secondly, the temporal encoder-decoder structure of VAE-RGOM and ATT-RGOM enables these models to learn a low-dimensional (latent space) manifold of the data. Ideally, this manifold untangles variation factors across distinct movements, clusters related motion descriptors, and aids in identifying joint dynamics across sequences. In the case of VAE-RGOM, it disentangles the dynamics and postures in terms of the ELBO. For ATT-RGOM, this latent space allows the attention mechanism to interpret the hidden mechanisms and connections underlying the motion descriptors sequences. According to the presented metrics, the encoding done by VAE-RGOM allowed the most accurate movement simulations. VAE-RGOM may outperform ATT-RGOM since it models a probability distribution over future postures rather than making point estimates. Particularly, VAE-RGOM yields the highest generation performance for movements from the datasets ERGD and TVA. These datasets are the largest ones and correspond to the simplest movements with low intraclass variability. In these, the movements were performed in a more controlled setting. For instance, in ERDG, the subjects performed diverse postures in a laboratory, receiving constant instructions on how to perform them. In the case of TVA, the operators were recorded in a production cell performing the same tasks repeatedly for several hours with slight variation between repeats. In addition, the movements in TVA primarily involved manipulating objects with their hands, in contrast to the movements performed, for instance, by the craftsmen and farmers (GLB, SLW, and MSC), who had to employ their entire bodies to perform their work properly, making the modeling and generation of these movements the hardest.

The most challenging movements to replicate were those associated with mastic cultivation. The reason could be a bias in the training data, as MSC was the smallest dataset and involved movements where the farmer most of the time moved while kneeling. When the farmer moved to reach the tree or objects, he usually repositioned the legs while kneeling to improve balance. In the other six datasets, the subjects were mostly standing while performing their tasks. This may have prevented the networks from fully learning the dynamics of the legs when they are flexed. 

\subsection{Tolerance intervals for evaluating movement performance}
Some applications of human movement analysis involve evaluating a subject's performance by analyzing the similarity between two movements. For instance, to examine gait \cite{Ezati2019} or to instruct proper tai-chi postures \cite{Liao2021}. Another application of measuring movement similarity is for communicating with a human-computer interface \cite{Caramiaux2015}. The motion representations presented in this paper can be used to compare distinct motions. One method is to compare their GOM mathematical representations directly; alternatively, tolerance intervals of the joints' motion can be calculated if the application calls for assessing how well a user can replicate a movement.

\begin{figure*}[]
\centering 
\sidesubfloat[]{%
  \includegraphics[width=0.4\columnwidth]{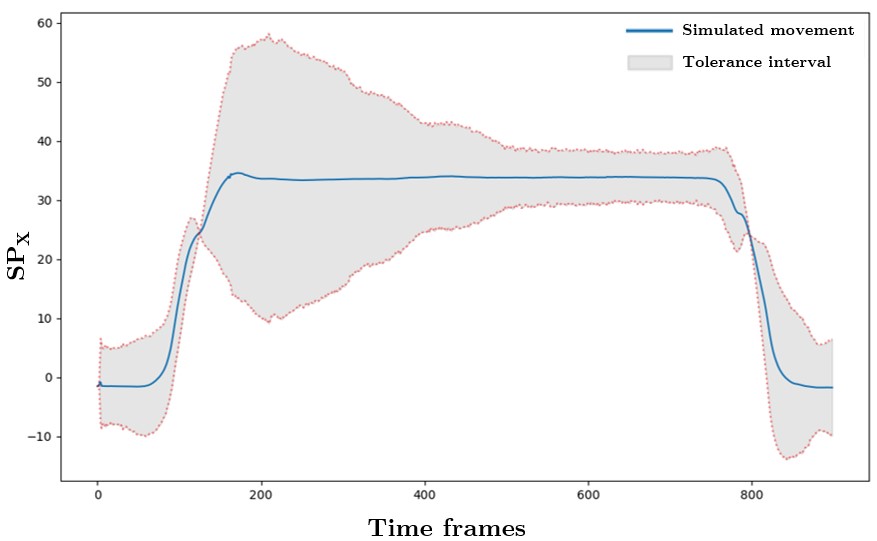}\label{fig:ti_a}
}
\sidesubfloat[]{%
  \includegraphics[width=0.4\columnwidth]{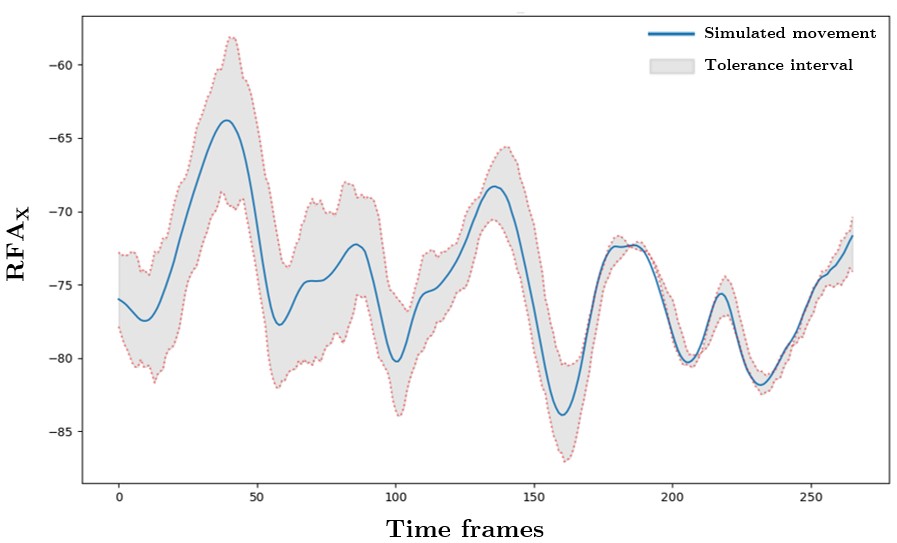}\label{fig:ti_b}
}
\caption{Tolerance intervals: \protect\subref{fig:ti_a} Shows the tolerance intervals for the movement of bending forward ($ERGD_7$); \protect\subref{fig:ti_b} Shows the tolerance intervals for the movement of embroidery ($MSC_5$).}\label{fig:ti}
\end{figure*}

These tolerance intervals vary throughout the time series and indicate the range of motion acceptable for properly executing a specific movement. To calculate the tolerance intervals, all repetitions of a movement are first aligned in time using DTW and a template movement. Then, their time-varying GOM representation is trained to extract their aligned coefficients. The tolerance intervals can then be defined using the standard distribution of the coefficients ($\sigma_{n,t}$) for each time step $t$:
\begin{align}\label{eq:ti1}
\mu_{n,t} &= \frac{\sum_{i=1}^{R}\alpha_{i,n,t}}{R} & \sigma_{n,t} &= \sqrt{\text{\footnotesize$\frac{\sum_{i=1}^{R}(\alpha_{i,n,t}-\mu_{n,t})^{2}}{R}$}}
\end{align}
where $R$ is the number of repetitions of a movement and $n$ the number of coefficients. One or two standard deviations can be defined for calculating the tolerance intervals for the correct execution of a movement. Figure \ref{fig:ti} illustrates two examples of tolerance intervals defined as two standard deviations.
\floatsetup[figure]{style=plain,subcapbesideposition=top}
The first example consists of the movement of bending forward more than $60^{\circ}$ ($\text{ERGD}_7$) for six seconds, and the second is the embroidery of a mastic tree ($\text{MSC}_5$), which consist of the farmer making incisions with a small knife to a mastic tree. In Figure \ref{fig:ti_a}, it can be observed that the tolerance interval is wider at the moment the subjects bend, as the subjects need to readjust their posture after bending to maintain balance and prevent falling forward. The second example illustrates that there is greater variation in the first cuts of the tree, which may be due to the fact that the harvester does not begin cutting at the same location. However, at the end of the movement, there is more precision in the cutting. Note that the tolerance intervals in the first example for $\text{ERGD}_7$ are wider than in the other, as they are generated using representations from several subjects. In contrast, the tolerance intervals for $\text{MSC}_5$ are calculated using motion models from a single farmer, which performed movements with higher precision.

\section{Conclusion}\label{sec:concl}
This work primarily focused on creating methodologies for training explainable human motion models. The applicability of state-space models for developing a generalized motion understanding framework was investigated. Consequently, three approaches that adhere to the structure of the Gesture Operational Model were proposed. The proposed methods were able to parameterize the conditional distributions specified in the state-space models. Also, they exhibited their potential to learn human movements in a general and scalable way, as they were able to fit data distributions from reduced data sets and recorded with different subjects in different scenarios. 

The trained models allowed the body dexterity analysis of industrial operators and skilled craftsmen, as well as the artificial generation of professional movements. Additionally, with the learned motion representations, it was possible to perform a selection of meaningful motion descriptors for modeling a set of human movements. This selection method could be utilized for a broad range of applications requiring modeling specific movements using a minimal sensor configuration. Determining the minimal motion descriptors to measure allows for the adoption of less invasive MoCap technologies, such as smartphones and smartwatches, that could also measure these motion descriptors. 

Depending on the nature of the motion-based application, either the statistical (KF-RGOM) or the deep-learning approaches (VAE-RGOM and ATT-RGOM) may be used to train the proposed human motion representations. For instance, whether single or multiple human movements are analyzed, as well as the processing power constraints. VAE-RGOM and ATT-RGOM are able to generate human movements more accurately and can be scaled to provide a greater variety of human motions. In addition, these approaches generate the representations of all full-body motion descriptors simultaneously, unlike KF-RGOM, which requires modeling one motion descriptor at a time, meaning training separately 57 models (one per descriptor) for simulating full-body movements. Nevertheless, the training of VAE-RGOM and ATT-RGOM is data-intensive, requiring a large volume of data and a significant amount of computing power. KF-RGOM, on the other hand, is sufficiently accurate to model specific human movements using one-shot training, which demands less computational power than deep-learning methods.

As potential future work, the proposed analytical models can be integrated with neurophysiological techniques that, for example, account for muscle activity and motor cortex activity. This combination would allow for a more thorough approach that could provide a neurophysiological roadmap of complex body dexterity. However, because of the inherent complexity and the sheer amount of data it requires, such a complete study that considers all these neurophysiological factors has not been done to this point.

%% Loading bibliography style file
\bibliographystyle{elsarticle-num}
%\bibliographystyle{cas-model2-names}

% Loading bibliography database
\bibliography{refs}

\end{document}